\documentclass[11pt]{article}

\usepackage[preprint]{acl}

\usepackage{times}
\usepackage{latexsym}

\usepackage[T1]{fontenc}
\usepackage[utf8]{inputenc}

\usepackage{microtype}

\usepackage{inconsolata}

\usepackage{graphicx}

\usepackage[most]{tcolorbox}

\setcitestyle{numbers,square}

\usepackage{booktabs}
\usepackage{tcolorbox}
\usepackage{multirow}
\usepackage{amsmath}
\usepackage{xcolor}           % make sure xcolor is loaded
\definecolor{shadecolor}{rgb}{0.9,0.9,0.9}  % light gray

\usepackage{tikz}
\usetikzlibrary{arrows.meta, positioning}
\usepackage{amssymb}

\usepackage{longtable}

\usepackage{enumitem}

\usepackage{graphicx} 
\usepackage{subcaption}

\usepackage{listings}
\title{How and Why LLMs Generalize: A Fine-Grained Analysis of LLM Reasoning from Cognitive Behaviors to Low-Level Patterns}

\author{
  \textbf{Haoyue Bai\textsuperscript{1}},
  \textbf{Yiyou Sun\textsuperscript{2}},
  \textbf{Wenjie Hu\textsuperscript{1}},
  \textbf{Shi Qiu\textsuperscript{2}},
\\
  \textbf{Maggie Huan\textsuperscript{3}},
  \textbf{Peiyang Song\textsuperscript{4}},
  \textbf{Robert Nowak\textsuperscript{1}},
  \textbf{Dawn Song \textsuperscript{2}},
\\
  \textsuperscript{1}University of Wisconsin, Madison,
  \textsuperscript{2}University of California, Berkeley,
  \\
  \textsuperscript{3}University of Pennsylvania,
  \textsuperscript{4}California Institute of Technology,
}

\begin{document}
\maketitle
\begin{abstract}
Large Language Models (LLMs) display strikingly different generalization behaviors: supervised fine-tuning (SFT) often narrows capability, whereas reinforcement-learning (RL) tuning tends to preserve it. The reasons behind this divergence remain unclear, as prior studies have largely relied on coarse accuracy metrics. We address this gap by introducing a novel benchmark that decomposes reasoning into atomic core skills such as calculation, fact retrieval, simulation, enumeration, and diagnostic, providing a concrete framework for addressing the fundamental question of what constitutes reasoning in LLMs. By isolating and measuring these core skills, the benchmark offers a more granular view of how specific cognitive abilities emerge, transfer, and sometimes collapse during post-training. Combined with analyses of low-level statistical patterns such as distributional divergence and parameter statistics, it enables a fine-grained study of how generalization evolves under SFT and RL across mathematical, scientific reasoning, and non-reasoning tasks. Our meta-probing framework tracks model behavior at different training stages and reveals that RL-tuned models maintain more stable behavioral profiles and resist collapse in reasoning skills, whereas SFT models exhibit sharper drift and overfit to surface patterns. This work provides new insights into the nature of reasoning in LLMs and points toward principles for designing training strategies that foster broad, robust generalization.

\end{abstract}

\section{Introduction}

LLMs fine-tuned with long Chain-of-Thought (CoT) reasoning, DeepSeek-R1~\cite{guo2025deepseek}, OpenAI-o1~\cite{openai2024learning}, Claude-Sonnet~\cite{claude}, achieve strong results on math and science benchmarks, yet their ability to \emph{generalize} remains fragile. A notable pattern has emerged: models trained with supervised fine-tuning (SFT) often narrow their capability and overfit to surface patterns~\cite{sun2025climbing}, while those tuned with reinforcement learning (RL) better preserve or even enhance generalization~\cite{huan2025does, sun2025delta}. The underlying reasons remain unclear, partly because prior studies rely on coarse metrics such as end-to-end accuracy or pass@k, which obscure the behavioral dynamics that drive reasoning success or failure~\cite{wei2022chain, openai2023gpt4}.

We argue that reasoning is not a monolithic property but rather an emergent composition of \emph{atomic core skills}, calculation, simulation, fact retrieval, enumeration, and diagnostic self-checking, that interact with statistical patterns acquired during training~\cite{kuang2025scores,niu2024large}. Accuracy on a final answer can mask weaknesses in these intermediate skills: a model may recall the correct formula yet incorrectly simulate a process, or arrive at the right solution through superficial pattern matching. Understanding how post-training reshapes these skills is crucial for explaining divergent generalization trends.

\begin{figure*}[t]
\centering
  \includegraphics[width=0.65\textwidth]{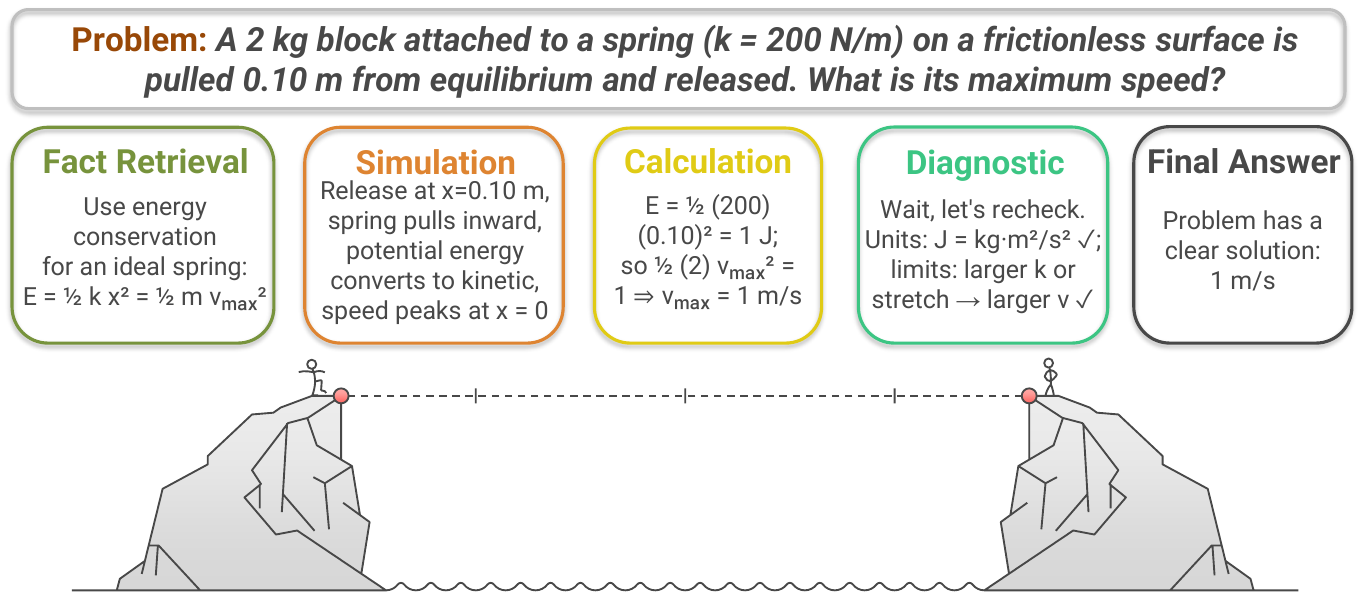}
  \caption{Decomposing reasoning into atomic cognitive skills. A spring–block example is solved step by step: fact retrieval, simulation, calculation, and diagnostic, checking each function as an isolated cognitive skill. When combined, these skills form a coherent reasoning trace that enables solving real-world problems, illustrating how complex reasoning emerges from the composition of simpler, specialized components.}
  \label{fig:reasoning_springblock}
\end{figure*}

\paragraph{Illustrative Example.}  
Figure~\ref{fig:reasoning_springblock} shows a classic spring block problem: a 2\,kg block attached to a spring is pulled 0.10\,m from equilibrium and released. Solving it requires sequential use of several atomic core skills. The solver first \textbf{retrieves} the conservation of energy relation for a spring, then \textbf{simulates} the block’s motion as potential energy converts to kinetic, \textbf{calculates} the peak speed at the equilibrium point, and finally performs a \textbf{diagnostic} of units and boundary conditions. Each skill plays a distinct functional role; failure at any stage breaks the reasoning chain. While not needed in this particular example, a separate \textbf{enumeration} skill, systematically listing possible cases or configurations, is equally crucial for comprehensive reasoning, especially in combinatorial or case-based problem settings. This illustrates that what we call “reasoning” in practice is the coordinated composition of such simpler cognitive elements.

Existing benchmarks and analyses fall short of capturing this structure. Large mixed-domain corpora such as \textit{Numina-Math}~\cite{numina_math_datasets} and \textit{Omni-Math}~\cite{gao2024omni} obscure which skills drive success; controlled datasets like \textit{GSM-Symbolic}~\cite{mirzadeh2024gsm} and \textit{GSM-PLUS}~\cite{li2024gsm} target narrow sub-skills; and conventional accuracy metrics hide shifts in intermediate behavior~\cite{cobbe2021training}. To fill this gap, we introduce a novel benchmark that explicitly decomposes reasoning into atomic core skills across mathematics, scientific reasoning, coding, and non-reasoning tasks, complemented by probes for low-level statistical patterns such as distributional divergence and shifts in word-frequency profiles. This design enables us to track how individual skills and statistical tendencies evolve through the course of post-training.

Our empirical results reveal several findings:
(a) \textbf{RL preserves balanced cognitive skills.}
RL-tuned models maintain rounded radar profiles across core skills: calculation, enumeration, simulation, fact retrieval, and diagnostic, indicating more stable and broad-spectrum reasoning. This balance persists across both in-domain (math) and out-of-domain (physics, non-reasoning) settings.
(b) \textbf{SFT induces over-specialization and drift.}
SFT-tuned models display jagged radar shapes, with strong spikes in a narrow skill (often diagnostic or calculation) but dips below baseline in others, such as simulation and enumeration. This pattern signals over-fitting to surface heuristics and loss of transferable reasoning capacity.
(c) \textbf{Training objective matters more than parameter magnitude.}
The observed divergence arises despite SFT and RL modifying a comparable proportion of model parameters. The difference stems from their optimization objectives, which shape how cognitive skills are retained or distorted during post-training.

Our findings provide a behavioral account of why post-training strategies diverge in their generalization effects and highlight the importance of strengthening atomic core skills as a foundation for robust and interpretable reasoning.

\section{Related Works}

\textbf{Post-training for Reasoning in LLMs.}
Recent progress in large language models has underscored the importance of specialized post-training strategies, especially fine-tuning for reasoning \cite{ziegler2019fine, sun2025climbing}.  
Chain-of-thought (CoT) prompting introduced by \cite{wei2022chain} encourages step-by-step explanation and significantly improves performance on symbolic and multi-step reasoning tasks \cite{longpre2023flan,lambert2024tulu}.  
Recent extensions such as DeepSeek-R1 (Team, 2025a) combine CoT with reinforcement-learning-based optimization to further enhance reasoning and have achieved state-of-the-art results on math, logic, and competitive programming benchmarks \cite{hendrycks2021measuring,guo2025deepseek}.

Post-training methods for reasoning typically fall into two categories: supervised fine-tuning (SFT) and reinforcement-learning-based tuning (RL) \cite{ziegler2019fine, chu2025sft}.  
SFT trains models to replicate explicit reasoning traces collected from annotated solutions \cite{wang2022self,wei2022chain}, whereas RL rewards models for accurate and logically coherent reasoning without requiring explicit intermediate supervision \cite{chu2025sft, ziegler2019fine}.  
This difference in optimization objective leads to distinct generalization patterns: SFT often narrows the behavioral diversity of models and overfits to surface heuristics \cite{perez2021true}, whereas RL tends to preserve or even enhance broader reasoning capacities but can introduce reward hacking and biases \cite{ouyang2022training,bai2022constitutional}.
Despite these insights, most prior analyses rely primarily on coarse outcome-based metrics such as final-task accuracy, leaving the process-level and representation-level dynamics of post-training effects underexplored.

\textbf{Generalization and cross-domain reasoning.} Generalization to tasks or domains outside the training distribution, remains a central challenge for LLMs.
This challenge has been extensively studied in discriminative models, where systems often fail under covariate, diversity, or semantic shifts despite strong in-distribution performance \cite{bai2021decaug,bai2021ood,ye2022ood,baihypo,bai2024aha}.
While scaling laws highlight global performance trends across model size and data \cite{kaplan2020scaling,hoffmann2022training}, fine-tuning often induces qualitative shifts in reasoning robustness and error modes \cite{zhang2021understanding,hendrycks2021measuring}.
Comparative studies suggest that SFT-dominated reasoning models frequently over-specialize, losing robustness to new task formats or domains, whereas RL-based fine-tuning helps retain transferable skills \cite{openai2023gpt4}.
For example, OpenAI’s o1 model excels in STEM reasoning but has raised concerns about versatility on other tasks \cite{jaech2024openai}. 
Recent evaluations emphasize that reasoning-oriented fine-tuning can improve performance on target tasks but sometimes comes at the cost of cross-domain generalization \cite{sun2025omega, cheng2025revisiting}. Our work complements these studies by moving beyond aggregate accuracy to analyze reasoning at a finer granularity. We decompose reasoning into core cognitive behaviors: calculation, simulation, enumeration, fact retrieval, and diagnostic, and reveal how SFT and RL tuning differentially shape these components, providing a clearer account of divergent generalization effects.

\textbf{Cognitive behaviors and representation-level shifts.}
Accuracy alone can hide weaknesses in intermediate reasoning. We therefore decompose performance into five measurable behaviors, \emph{calculation}, \emph{simulation}, \emph{fact retrieval}, \emph{enumeration}, and \emph{diagnostic checking}, making these skills explicit and comparable across domains and training stages. While BIG-bench, MATH, and related psychometric work touch on similar abilities \citep{srivastava2022beyond,hendrycks2021measuring,lampinen2022can}, our framework operationalizes them with targeted prompts and metrics; see also domain-specific analyses such as \emph{From Scores to Skills} in finance \citep{kuang2025scores} and broader links to cognitive science \citep{niu2024large}.

Fine-tuning not only changes outward behavior but also the internal representations that support it. CoT can elicit more systematic reasoning \citep{wei2022chain}, yet traces may be unfaithful to internal computations \citep{turpin2023language}. Sparse autoencoders and activation steering reveal interpretable subspaces tied to reasoning features \citep{hookedsae2024,turner2023activation}. By jointly measuring behavioral skills and representational structure, we characterize how SFT and RL differentially shape both the internal and external facets of LLM reasoning.

\begin{figure}[t]
\centering
  \includegraphics[width=0.45\textwidth]{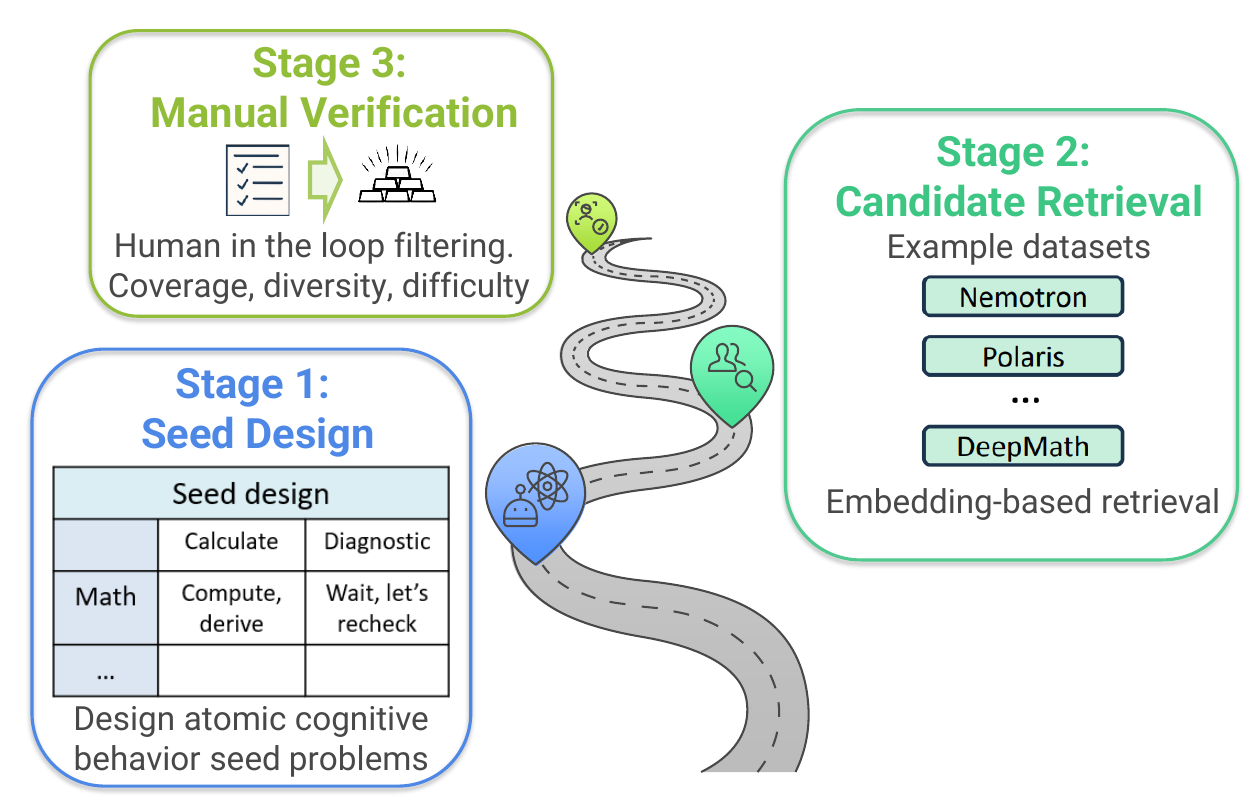}
  \caption{Overview of our three-stage benchmark construction pipeline. \textbf{(1) Seed Design} defines atomic seeds by domain and behavior; \textbf{(2) Candidate Retrieval} expands them via embedding search over public datasets; \textbf{(3) Manual Verification} filters for coverage, diversity, and difficulty, yielding a curated benchmark for fine-grained reasoning analysis.}
  \label{fig:pipeline}
\end{figure}

\begin{table*}[htbp]
\centering
\caption{\footnotesize Example problem templates across four domains and five cognitive behaviors. Representative tasks illustrate the diversity of problem types used to evaluate model capabilities. }
\label{tab:problem_templates}
\scalebox{0.8}{
\begin{tabular}{@{} l p{3.5cm} p{3.5cm}  p{3.5cm}  p{3.5cm} @{}}
\toprule
\textbf{Category} & \textbf{Math Reasoning} & \textbf{Scientific Reasoning} & \textbf{Coding} & \textbf{Non-reasoning} \\
\midrule

\multirow{5}{*}{\textbf{Calculation}}
  & \textbf{Definition:} The execution of arithmetic, algebraic, or symbolic operations to derive quantitative or formal results from given inputs. 
  & \textbf{Classical mechanics:} A proton enters a uniform magnetic field perpendicularly, compute the radius of its circular path. 
  &\textbf{Matrix primitivity:} Given a non-negative matrix $A\in\mathbb{R}_{\ge 0}^{n\times n}$, decide whether there exists an integer $k \ge 1$ such that $A^{k} > 0$ 
  & N/A 
  \\
  & \textbf{Example:} Find the second largest prime factor of 49766. 
  & \textbf{Magnetism: } In an RC circuit, compute capacitor’s charge at a given time after the circuit is closed. 
  &\textbf{Unique array.}
Unique $s$, construct $a,b\ge0$, $a_{i}+b_{i}=s_{i}$, each unique after removing $\le\!\lfloor n/2\rfloor$ entries. &
\\
\midrule

\multirow{5}{*}{\textbf{Enumeration}}
  & \textbf{Definition: }The systematic and exhaustive generation of all elements, cases, or options satisfying a set of explicit constraints.            
  & \textbf{Spin Projection:} \( 8 \) non-interacting spin-\(\tfrac{1}{2}\) particles, find number of microstates with total spin projection \( M_{s} = +1 \).
  & \textbf{Set Partitions: }
  All partitions of $\{1..n\}$ in canonical order; consecutive partitions differ by moving one element between blocks. 
  & \textbf{Rewards program features enumeration: } Create a webpage section listing Eventsfy's Sparks Awards program features.  \\
  & \textbf{Example:} You have letters {‘q’: 4, ‘l’: 2, ‘b’: 2}. Distribute them into 3 labeled boxes of capacities [4, 2, 2]. How many ways? 
  & \textbf{Atomic-Orbital Counting:} 
The number of microstates for a \( d^{3} \) electron configuration consistent with the Pauli principle. 
& \textbf{Primes in Ranges: }
  For $t$ queries $[m,n]$ (with $n-m\le 10^5$), list all primes in each range. 
  & \textbf{Article structure enumeration: }Create academic article on integrating ASHA workers into geriatric care and NCD clinic limitations.  \\
\midrule

  \multirow{5}{*}{\textbf{Simulation}}
  & \textbf{Definition:} The symbolic enactment of a process, system, or sequence of operations to predict or trace its behavior over time.        
  & \textbf{Cooling Process:}  An object cools from \(90^\circ\mathrm{C}\) to ambient \(20^\circ\mathrm{C}\) Newton’s law; compute \(T\) after \(30\,\mathrm{min}\) with \(k = 0.02\,\mathrm{min^{-1}}\). 
  &\textbf{Robot Path with One Extra Move.}
Given a walk $s$ over $\{W,A,S,D\}$, , insert one move to minimize the bounding box area.
& \textbf{Weather Simulator: } A program forecasts weather and suggests clothing based on temperature and rain conditions. \\
  & \textbf{Example:} Integers \(1\!-\!120\) on board. 
Each minute replace \(n\) by \(d(n+3)\), divisor-count. Find total sum on board after 1 day.
  & \textbf{Decay Chain:} Nuclide \(A\) decays to \(B\) (half-life 2 h), then \(B\) decays to stable \(C\) (half-life 8 h); find the time when \(B\) peaks. 
  & \textbf{Furlo/Rublo Coin Game.}
Replace $x$ by $y\in[\lfloor x/4\rfloor,\lfloor x/2\rfloor]$;  
winner given by xor of Grundy $g(x)$, which periodic in $\log x$. 
& \textbf{Gaming-related simulation: } Creating a vodcast script analyzing how sound displays mood in the mobile game Pixel Gun 3D. \\
\midrule

  \multirow{5}{*}{\textbf{Fact Retrieval}}
  & \textbf{Definition: }(1) Pose the problem naming the required theorem.  
(2) Pose it without the theorem and check if the model applies it correctly.     
&\textbf{Cylindrical Capacitor:} 
  Recall the capacitance formula for a long cylindrical capacitor; compute \(C\) for 
  \(a = 1\,\mathrm{mm}\), \(b = 5\,\mathrm{mm}\), \(L = 0.5\,\mathrm{m}\).   
  & \textbf{Mixing Rule in the Minimum-Smoke: } When combining adjacent mixtures of colors a and b, what are the smoke produced and the resulting color?
  & \textbf{Cross-Cultural Information Retrieval: }Identify sources for an article on the Alcatraz occupation and Indigenous Peoples' Day. \\
  & \textbf{Example:} Surface distance on radius r sphere between ($p_1, q_1$), and ($p_2, q_2$). 
  &\textbf{Bragg’s Law:} 
  Recall \(2d\sin\theta = n\lambda\); find the smallest \(\theta\) for 
  \(d = 0.2\,\mathrm{nm}\), \(\lambda = 0.15\,\mathrm{nm}\).  
  & \textbf{GCD-Constrained BST:} What gcd condition must each edge between adjacent vertices satisfy? 
  & \textbf{Procedural Knowledge: }Prerequisites for AP Chemistry topics (stoichiometry, electrochemistry).  \\
\midrule

  \multirow{5}{*}{\textbf{Diagnostic}}
  & \textbf{Definition:} Diagnostic in LLMs often accompanied by explicit self-check (e.g., "wait", "let's recheck").      
  & \textbf{Pursuit of a fox by a hound:}
 Centripetal-acceleration perturbed with false small-deviation expansion.
  & \textbf{Uniqueness vs. Duplicates:} counting the same solution twice because it appears in different positions. 
& \textbf{Counter-Factual: }"Is tree planting an eco-social activity?" with perturbed response claiming it is not.\\
  & \textbf{Example: } $
f_n = \frac{n x_1 + n^2 x_2}{n x}
\quad$ is perturbed as $\quad
\frac{n x_1 - n^2 x_2}{n x}
$. 
& \textbf{Falling hinged rods:} Energy-conservation problem perturbed to suggest small-angle oscillation approach.
& \textbf{Optimal vs. Sub-optimal:} Picking the shortest edge next in a path problem that needs full DP.
& \textbf{Self-contradictory:} Document identifies "Villainous" as Cartoon Network series, then labels incorrectly. \\

\bottomrule
\end{tabular}}
\label{tab:taxonomy}
\end{table*}

\section{Decompose “Reasoning” into Fine-grained Cognitive Behaviors}

A central question of this study is how different post-training regimes, supervised fine-tuning (SFT) versus reinforcement-learning-based tuning (RL), reshape not only the overall accuracy of large language models (LLMs) but also the composition of their underlying reasoning skills. To answer this, we develop a controlled benchmark and a meta-analysis framework that reveals how individual cognitive behaviors evolve under these regimes.

Our approach proceeds in three stages (Figure~\ref{fig:pipeline}). First, we identify representative model families and collect checkpoints at multiple stages of SFT and RL. Second, we construct a structured benchmark that spans four domains (Mathematics, Scientific Reasoning, Coding, and Non-Reasoning QA) and targets five core cognitive behaviors: \emph{calculation}, \emph{enumeration}, \emph{simulation}, \emph{fact retrieval}, and \emph{diagnostic checking}. Third, we use behavior-focused probes to analyze how training dynamics affect the balance among these sub-skills and associated statistical patterns (e.g., distributional divergence). This controlled design moves beyond aggregate accuracy to trace how SFT and RL redistribute effort across fundamental behaviors, clarifying their distinct generalization profiles.

\subsection{Benchmark Construction}

\noindent\textbf{Principles.}
Each item targets a single atomic behavior (calculation, enumeration, simulation, fact retrieval, diagnostic) and is curated for \emph{skill isolation}, coverage, diversity, and calibrated difficulty. Implementation details (templates, keyword lists, perturbations, rubrics) appear in Appendix~\ref{app:keywords}.

\noindent\textbf{Pipeline.}
Figure~\ref{fig:pipeline} summarizes the process. 
%\begin{itemize}
\begin{enumerate}[itemsep=1pt, topsep=1.5pt, parsep=1pt]
\item \textbf{Stage 1: Seed Design} creates atomic seeds for each behavior–domain pair, avoiding multi-skill conflation. 
\item \textbf{Stage 2: Candidate Retrieval} performs embedding-based nearest-neighbor search over large open repositories (e.g., Nemotron~\cite{bercovich2025llama}, Polaris~\cite{Polaris2025}, DeepMath~\cite{he2025deepmath}) to harvest surface variants that instantiate the same behavior in diverse contexts. 
\item \textbf{Stage 3: Manual Verification} applies human-in-the-loop filtering for skill isolation, coverage, diversity, and difficulty. The result is a high-fidelity benchmark that decouples reasoning behaviors from surface task formats.
%\end{itemize}
\end{enumerate}

\noindent\textbf{Behavior Assembly.}
\emph{Calculation, Enumeration, Simulation}: template seeds plus keyword queries retrieve items with explicit single-behavior traces from open sources, followed by deduplication and removal of multi-skill bleed. 
\emph{Fact Retrieval}: select questions whose solution hinges on a named person, theorem, or definition. We use two modes: \emph{Guided} (theorem named in the prompt) and \emph{Unguided} (theorem not named; verify its use in the solution). 
\emph{Diagnostic}: sample questions and apply minimal perturbations to a valid trace (logical contradiction, dropped condition, counterfactual). The prompt includes the question and the perturbed trace; the model must detect and correct the error (see example in Appendix~\ref{app:diagnostic-example}).

\noindent\textbf{Difficulty.}
Items are bucketed as \emph{easy}, \emph{medium}, or \emph{hard} using: (i) lightweight heuristics (operand scale, step count, constraint breadth, perturbation subtlety); (ii) reference-model success rates; and (iii) human adjustment prioritizing skill purity.

\noindent\textbf{Prompting (Appendix).}
Prompts consist of a standardized instruction header, behavior-specific payload (e.g., theorem tag or perturbed trace), output format, and fixed decoding settings, which are summarized in Appendix~\ref{app:prompt}.

\noindent\textbf{Human Evaluation.}
We check coverage across (behavior, domain, difficulty) cells, diversity of forms and entities, and the validity of difficulty labels; disagreements are adjudicated.

\noindent\textbf{Metrics.}
The primary metric is \textbf{accuracy} under standardized decoding. 
For Calculation, Enumeration, and Simulation, we use an exact match with unit normalization. 
For Fact Retrieval, \emph{Guided} requires a correct answer and theorem-consistent steps; \emph{Unguided} requires a correct answer and verified theorem use. 
For Diagnostic, we require a correct answer and explicit identification of self-check.

\subsection{Behavior–Domain Grid}

To support interpretable analysis, we organize the benchmark as a two-dimensional grid crossing five cognitive behaviors with four domains (Table~\ref{tab:taxonomy}). For each behavior–domain cell, the table provides a concise behavior definition and, in most cases, two representative examples. Due to space constraints, the \emph{Math} column anchors the behavior definitions and includes only one representative example per behavior; additional mathematical examples are provided in Appendix~\ref{app:math-addition}, Table~\ref{tab:math-examples}.

The five behaviors capture complementary facets of reasoning: 
(1) Calculation: quantitative manipulation of explicit formulas or equations; 
(2) Enumeration: systematic generation of combinatorial possibilities; 
(3) Simulation: mental or symbolic rollout of dynamics; 
(4) Fact Retrieval: access to stored knowledge such as definitions or constants; and 
(5) Diagnostic: identifying and correcting faulty reasoning or self-contradictions.

For example, as shown in Table~\ref{tab:taxonomy}, calculation tasks range from prime-factor queries in mathematics to computing capacitor charge in physics; enumeration spans classical combinatorics to set partitioning in code; simulation includes Newtonian cooling and other dynamical rollouts; fact-retrieval probes recall of theorems or scientific laws; and diagnostic checking tests recognition of flawed proofs or perturbed reasoning templates. This grid yields comprehensive yet disentangled coverage of reasoning behaviors, enabling us to track how SFT and RL affect each component skill within and across domains.

\begin{figure*}[t]
\centering
\begin{subfigure}[b]{0.245\textwidth}
  \includegraphics[width=\linewidth]{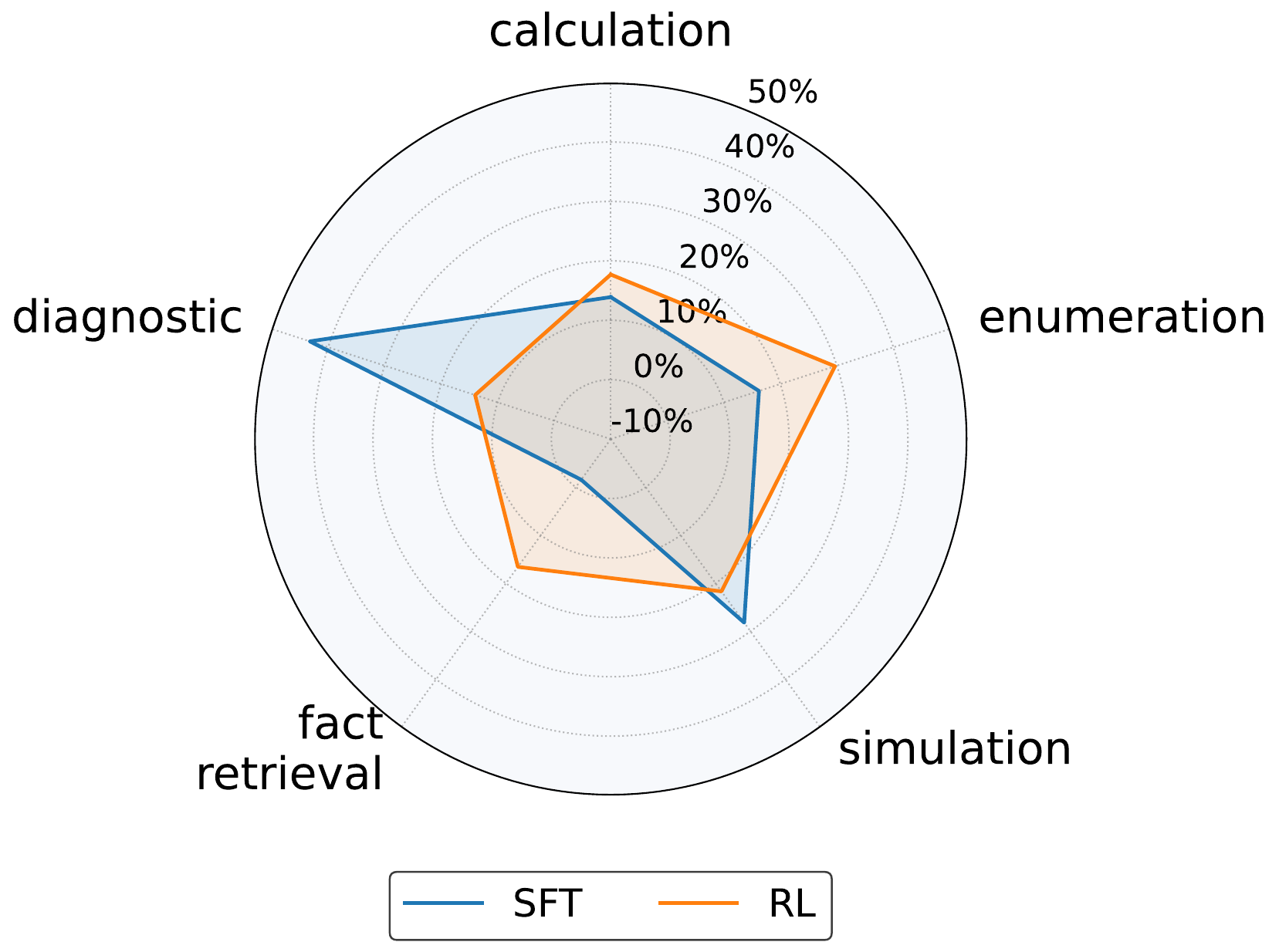}
  \caption{Math (14B)}
  \label{fig:sub1}
\end{subfigure}
\hfill
\begin{subfigure}[b]{0.245\textwidth}
  \includegraphics[width=\linewidth]{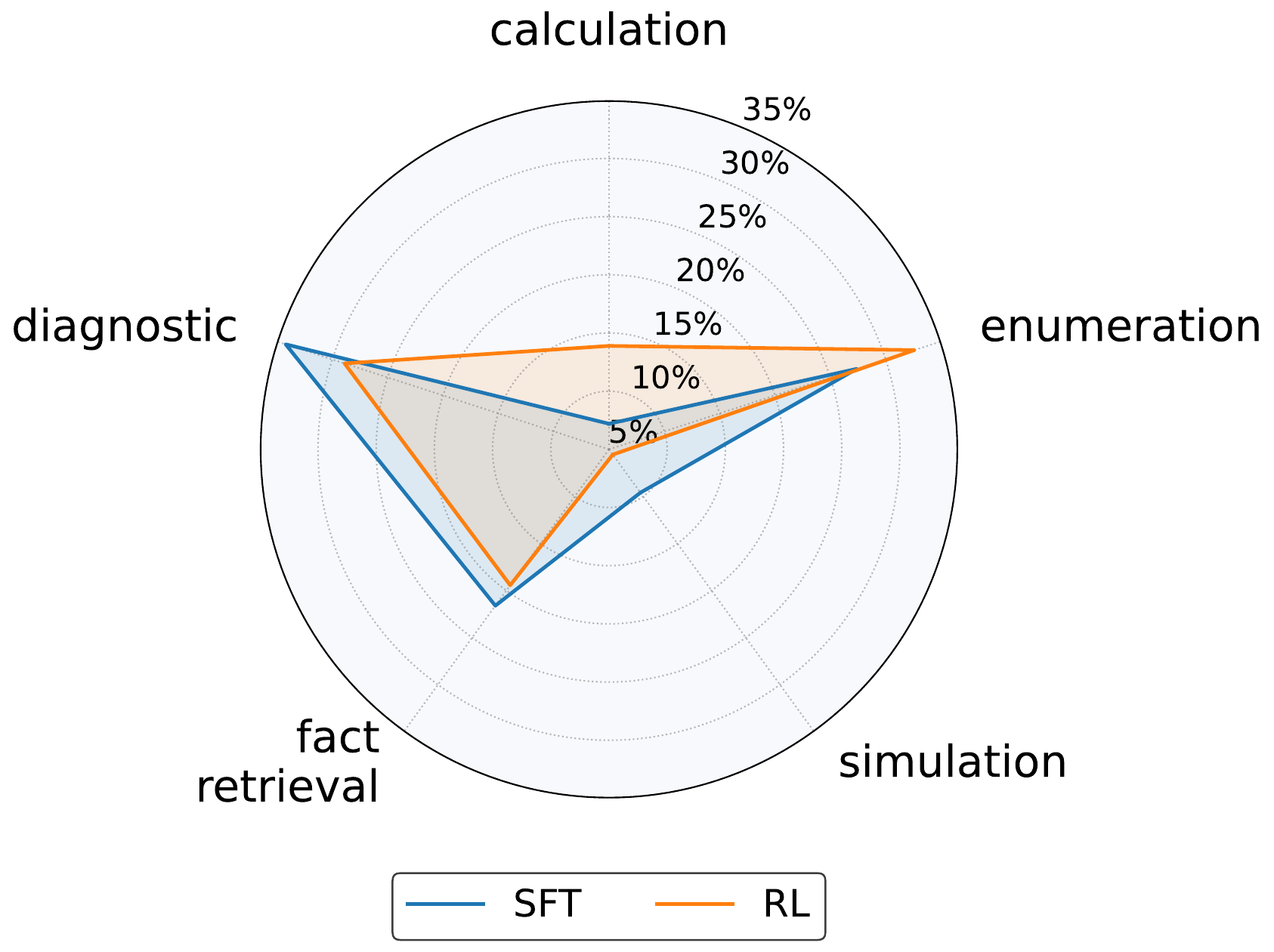}
  \caption{Physics (14B)}
  \label{fig:sub2}
\end{subfigure}
\hfill
\begin{subfigure}[b]{0.245\textwidth}
  \includegraphics[width=\linewidth]{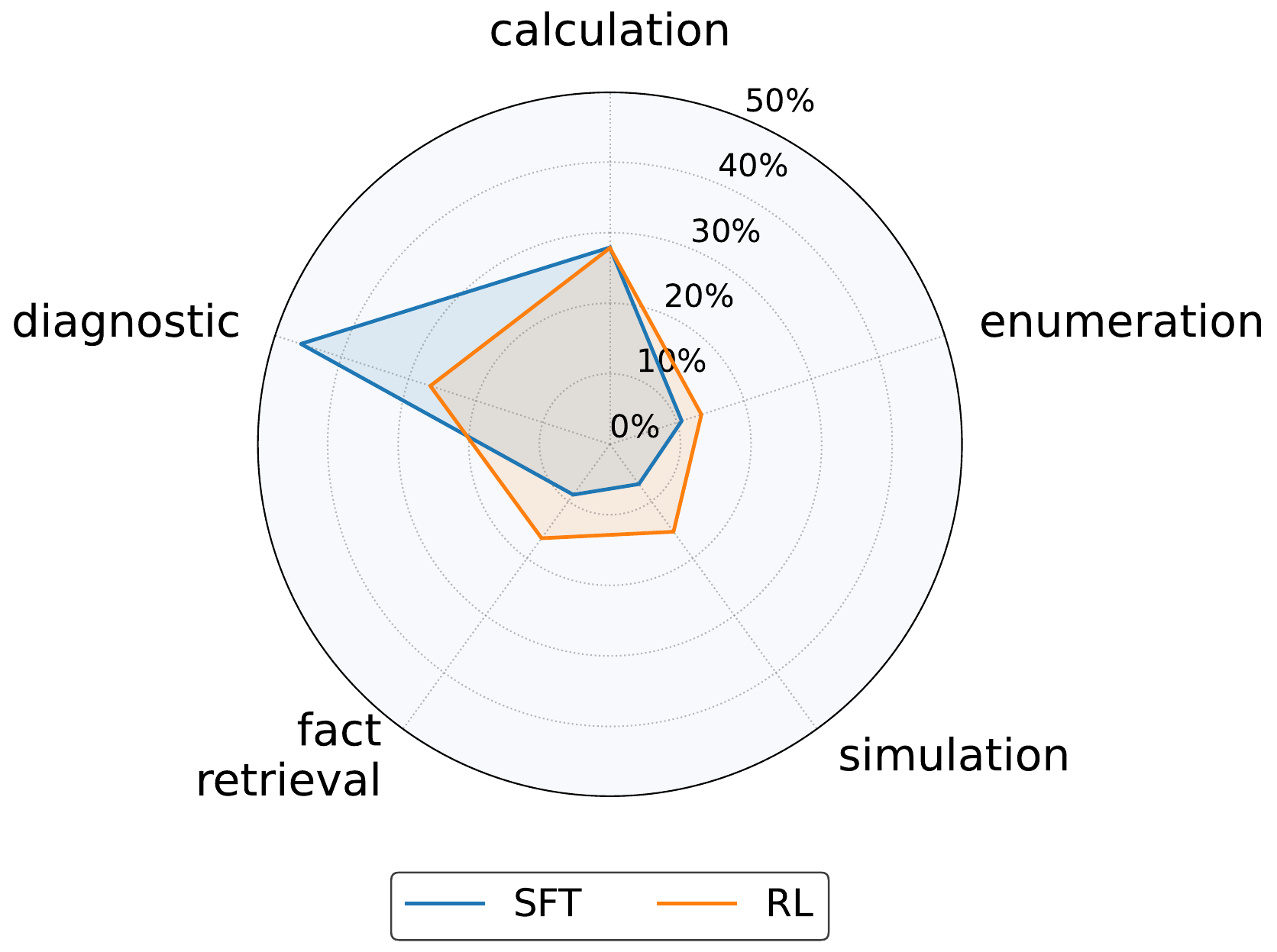}
  \caption{Math (4B)}
  \label{fig:sub3}
\end{subfigure}
\hfill
\begin{subfigure}[b]{0.245\textwidth}
  \includegraphics[width=\linewidth]{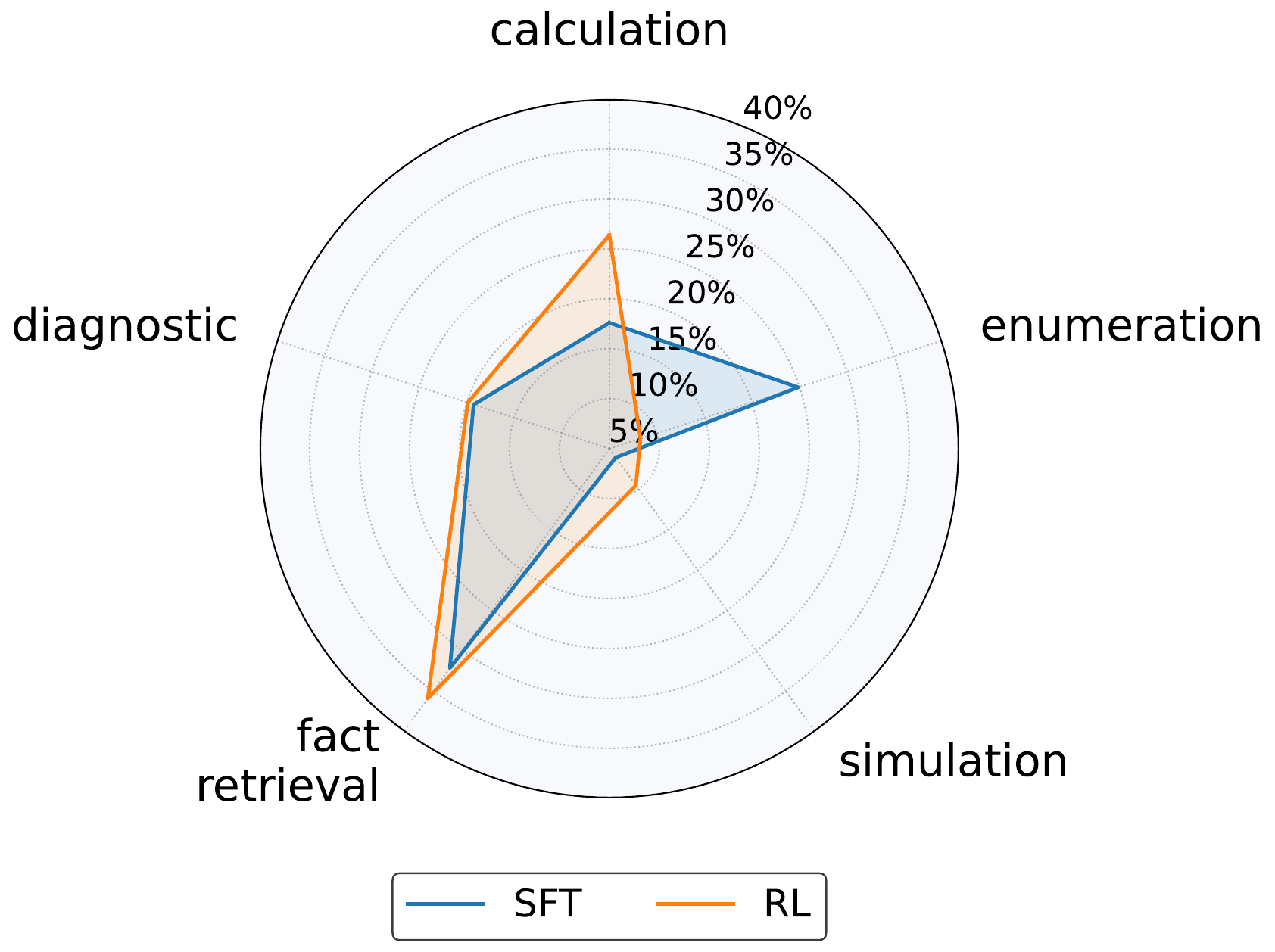}
  \caption{Physics (4B)}
  \label{fig:sub4}
\end{subfigure}
\caption{Radar plots of five cognitive behaviors (calculation, enumeration, simulation, fact retrieval, diagnostic checking). Each panel contrasts SFT (blue) vs. RL (orange) for (a) Math-14B, (b) Physics-14B, (c) Math-4B, and (d) Physics-4B. RL is more balanced; SFT often concentrates on a few skills.}
\label{fig:main_radar}
\end{figure*}

\section{Experiments}

\noindent\textbf{Experimental setup.}
We evaluate Qwen3-14B-Base and its SFT- and RL-tuned variants, along with smaller Qwen3-4-Base and Qwen3-1.7B-Base counterparts. To analyze training dynamics, we evaluate intermediate checkpoints throughout training. Following models used in \cite{huan2025does}, the RL model uses the Verl framework~\cite{sheng2025hybridflow} with a GRPO setup~\cite{shao2024deepseekmath} for RL fine-tuning, optimizing for answer-correctness rewards. The SFT model uses LLaMA-Factory~\cite{zheng2024llamafactory} to train on teacher-generated chain-of-thought traces through reject sampling.  
Evaluation is conducted on the above cognitive skills benchmark using accuracy. Detailed subcategory distributions for the proposed benchmark are provided in Appendix~\ref{app:data-distribution}. More details about training datasets, baseline models, and hyperparameters can be found in Appendix~\ref{app:expe-detail}.

\subsection{Cognitive-skill Profiles}

\begin{figure}[t]
\centering
\begin{subfigure}[b]{0.235\textwidth}
  \includegraphics[width=\linewidth]{14B_math.pdf}
  \caption{Math Domain}
  \label{fig:sub1}
\end{subfigure}
\hfill
\begin{subfigure}[b]{0.235\textwidth}
  \includegraphics[width=\linewidth]{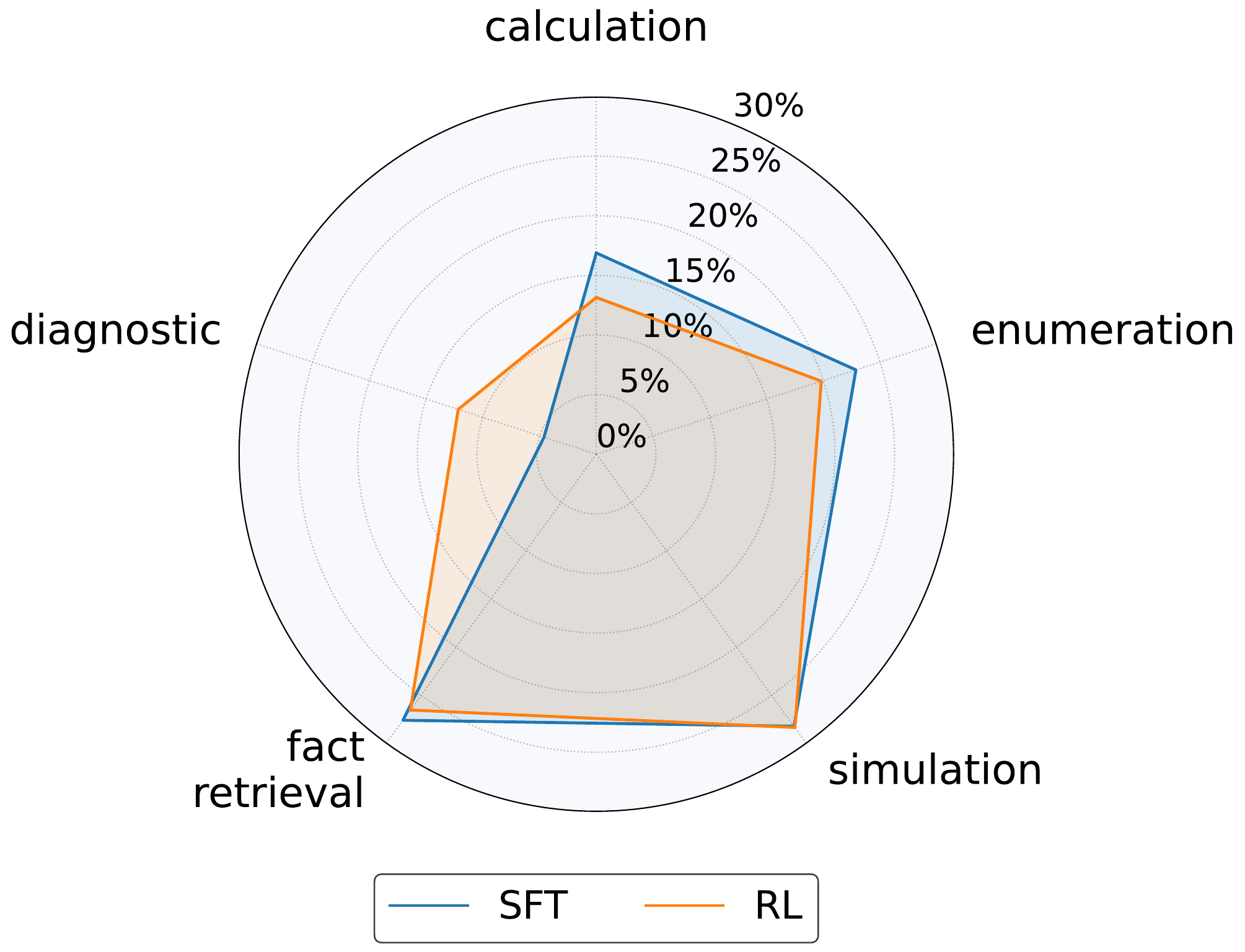}
  \caption{Non-reasoning Domain}
  \label{fig:sub2}
\end{subfigure}
\caption{Radar plots of post-training effects on math (a) and non-reasoning (b) using math-trained models. Both transfer weakly to non-reasoning; RL preserves a more balanced skill profile, whereas SFT shows uneven gains/losses out of domain.}
\label{fig:non_reasoning}
\end{figure}

\textbf{RL yields balanced skills and SFT over-specializes.} A striking pattern in Figure~\ref{fig:main_radar} is the irregular, jagged shape of the SFT curves compared to the rounded RL curves. The SFT-tuned models consistently exhibit asymmetric profiles, often showing pronounced spikes in a single skill (most commonly diagnostic) while dipping well below baseline in others, such as enumeration and simulation. This shape reflects a tendency toward over-specialization or over-fitting, where the model exploits superficial patterns in a limited subset of reasoning behaviors rather than maintaining a balanced skill set. For example, in Math-4B, SFT shows a sharp spike on diagnostic (about threefold above baseline), while enumeration and simulation fall below baseline; in Math-14B, it drops below baseline for calculation, simulation, and especially fact retrieval. In contrast, the RL-tuned profiles remain rounded and smoother in performance across all five cognitive behaviors, indicating that reinforcement-learning post-training promotes more even retention of diverse reasoning skills. This shape contrast is informative even when raw accuracy gaps between SFT and RL are modest, emphasizing that the training regime, not merely model scale or base capability, drives these differences in cognitive skill balance. In short, the rounded RL shapes highlight stable, broad-spectrum skill preservation, while the jagged SFT shapes expose imbalanced gains and losses, a visual signature of over-specialization that undermines transfer and generalization.

\begin{figure}[t]
\centering
\begin{subfigure}[b]{0.235\textwidth}
  \includegraphics[width=\linewidth]{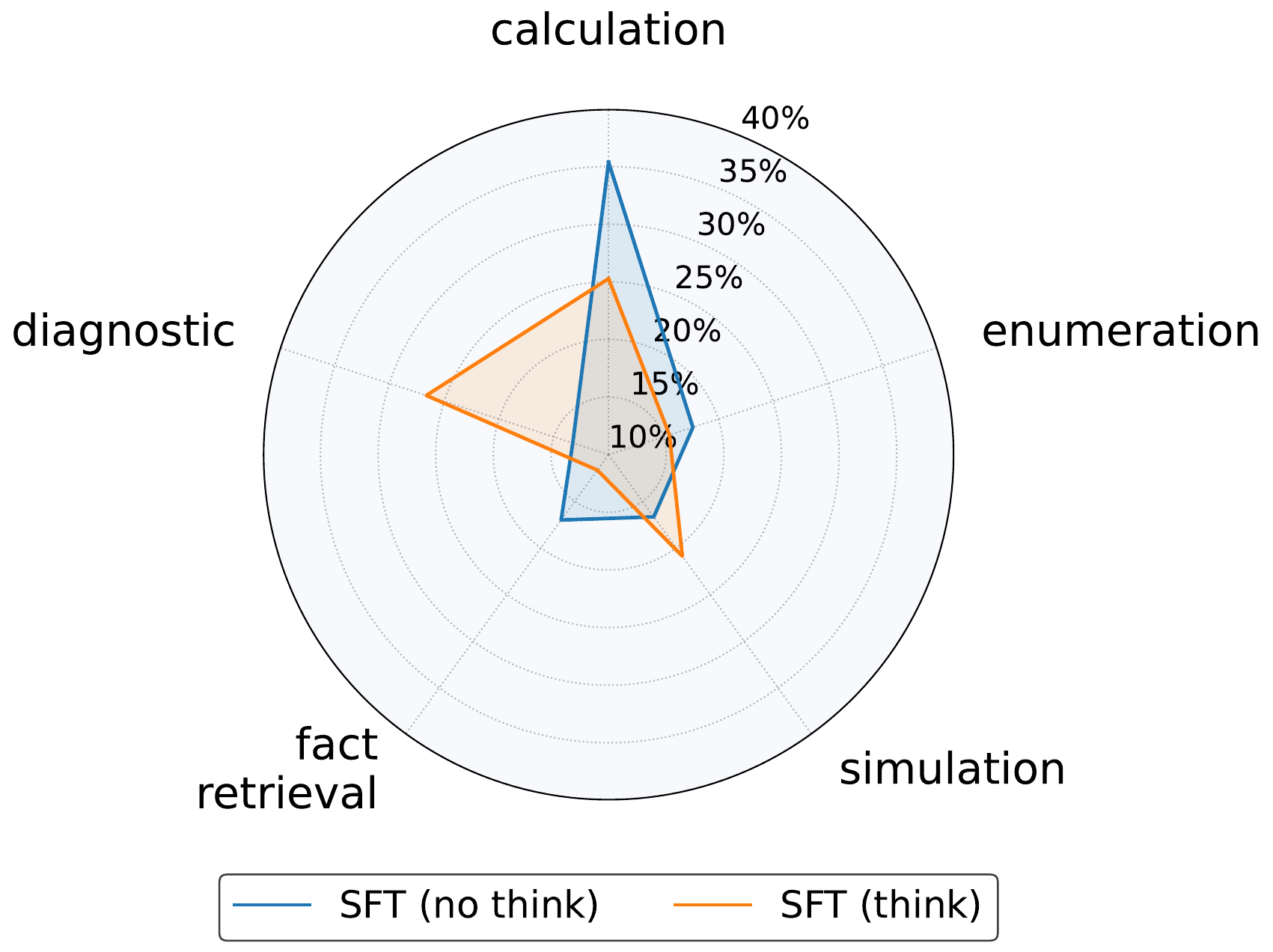}
  \caption{Math (SFT think vs no think)}
  \label{fig:sub1}
\end{subfigure}
\hfill
\begin{subfigure}[b]{0.235\textwidth}
  \includegraphics[width=\linewidth]{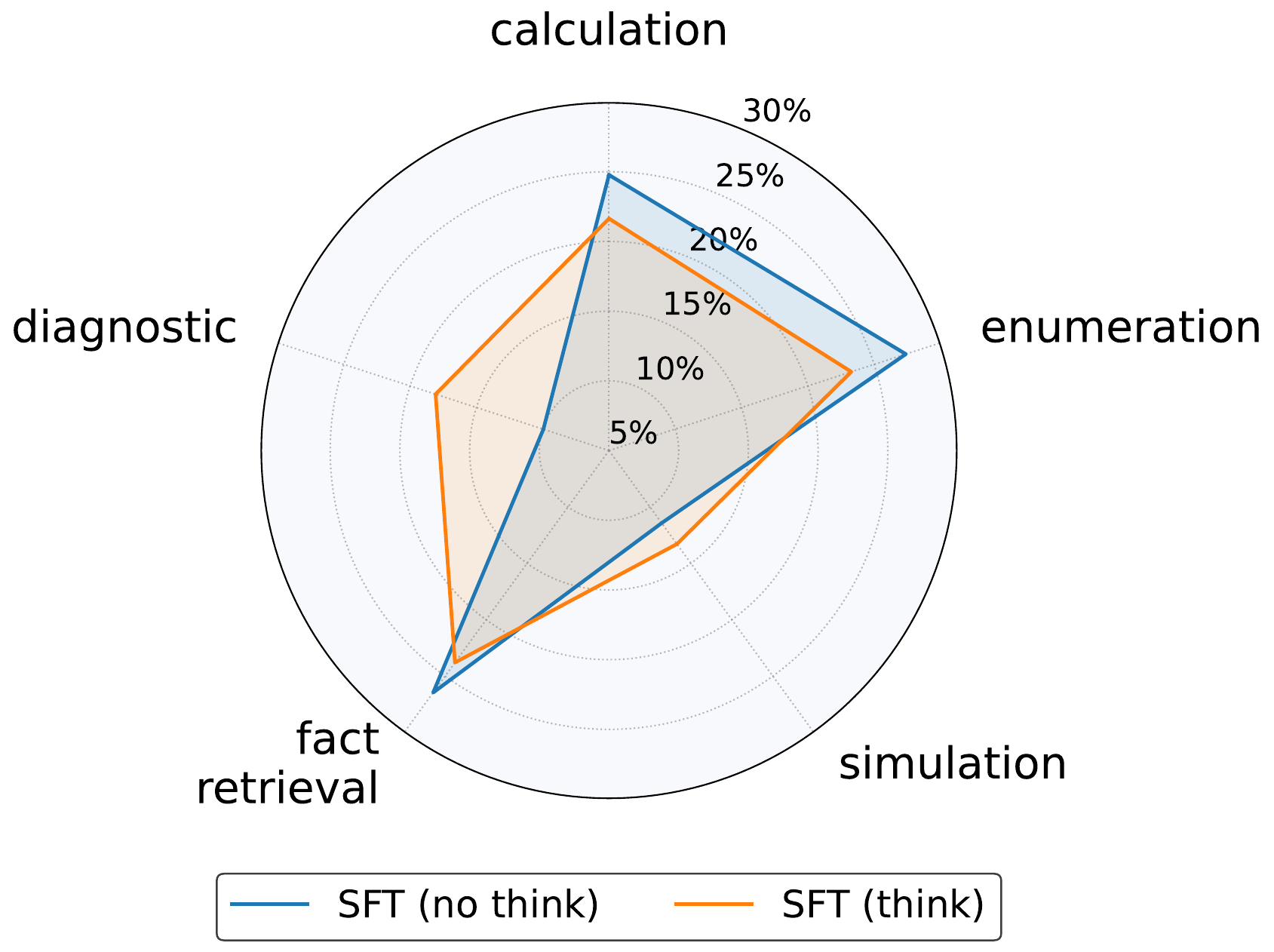}
  \caption{Physics (SFT think vs no think)}
  \label{fig:sub2}
\end{subfigure}
\caption{Shape comparison of SFT profiles with thinking (orange) vs no thinking (blue) on (a) Math and (b) Physics. Compared to the think model, the no-think model is even more spiky and calculation-dominated. All models are derived from the Qwen3-14B base; SFT and RL share the same base.}
\label{fig:think}
\end{figure}

\noindent\textbf{Long CoT rebalances SFT toward systematic reasoning.} Prior work shows SFT with teachers in thinking mode producing long CoT data generally outperforms SFT with teachers in non-thinking mode generating short CoT data \cite{huan2025does}. Figure~\ref{fig:think} compares SFT skill profiles under think vs.\ no-think mode for Math (a) and Physics (b). The no-think profile (blue) spikes on calculation while lagging on simulation, diagnostic checking, and enumeration. In contrast, the think models trained with long CoT (orange) are more rounded, redistributing capacity toward complementary processes essential for multi-step reasoning. Math is the in-domain setting; Physics is out-of-domain and shows transferability. The effect is especially clear in Physics, where long CoT patterns during post-training reduce calculation dominance and strengthen other components. Overall, long CoT training changes not only absolute scores but also the geometry of the skill profile, yielding a more systematic and integrated problem-solving approach. It provides richer supervision and encourages the model to internalize how to reason rather than merely what to answer. As a result, the learned routines transfer across domains, whereas short-CoT training, while often sufficient for in-domain factual recall, does not elicit robust, transferable multi-step reasoning.

\subsection{Cross-Domain Generalization}

\textbf{RL preserves a more balanced skill mix and degrades less than SFT under cross-domain transfer.}
Prior work \cite{huan2025does} shows that both RL and SFT degrade when transferring from math to non-reasoning domains, though RL typically degrades less and preserves a more balanced skill mix. Figure~\ref{fig:non_reasoning} compares post-training effects of SFT and RL on models trained only on math and evaluated on both math-specific and non-reasoning behaviors. The RL-tuned model displays a smoother, more balanced radar profile in the non-reasoning setting, whereas SFT is irregular, with spikes in a few skills and drops below baseline in others. This pattern suggests that RL post-training acts as a regularizer that preserves general reasoning competencies, while SFT overfits to math-specific patterns at the expense of cross-domain transfer. The relatively flat RL curve implies less skill interference and better retention of broad cognitive behaviors, even though absolute gains outside the math domain remain limited.

\subsection{Linking Behaviors to Sparse Latent Features}

\textbf{Shallow layers are near-Gaussian; deeper layers become heavier-tailed across base, SFT, and RL.} We analyze joint skewness–kurtosis of SAE latent weight distributions to characterize how training regimes reshape higher-order structure. In shallow layers, the base model is approximately symmetric and near-Gaussian, indicating relatively homogeneous latent activations. SFT induces mild shifts, slightly higher skewness and kurtosis, consistent with modest sharpening of latent selectivity. Across depth, all three models trend toward heavier-tailed, non-Gaussian behavior, reflecting increasing feature specialization. RL slightly amplifies this trend, with broader skewness–kurtosis dispersion and localized higher-variance regions, but not a qualitatively different regime. Overall, RL modestly increases heterogeneity and sparsity while preserving the depth-dependent progression already present in the base and SFT models.

\begin{figure}[t]
\centering
  \includegraphics[width=0.49\textwidth]{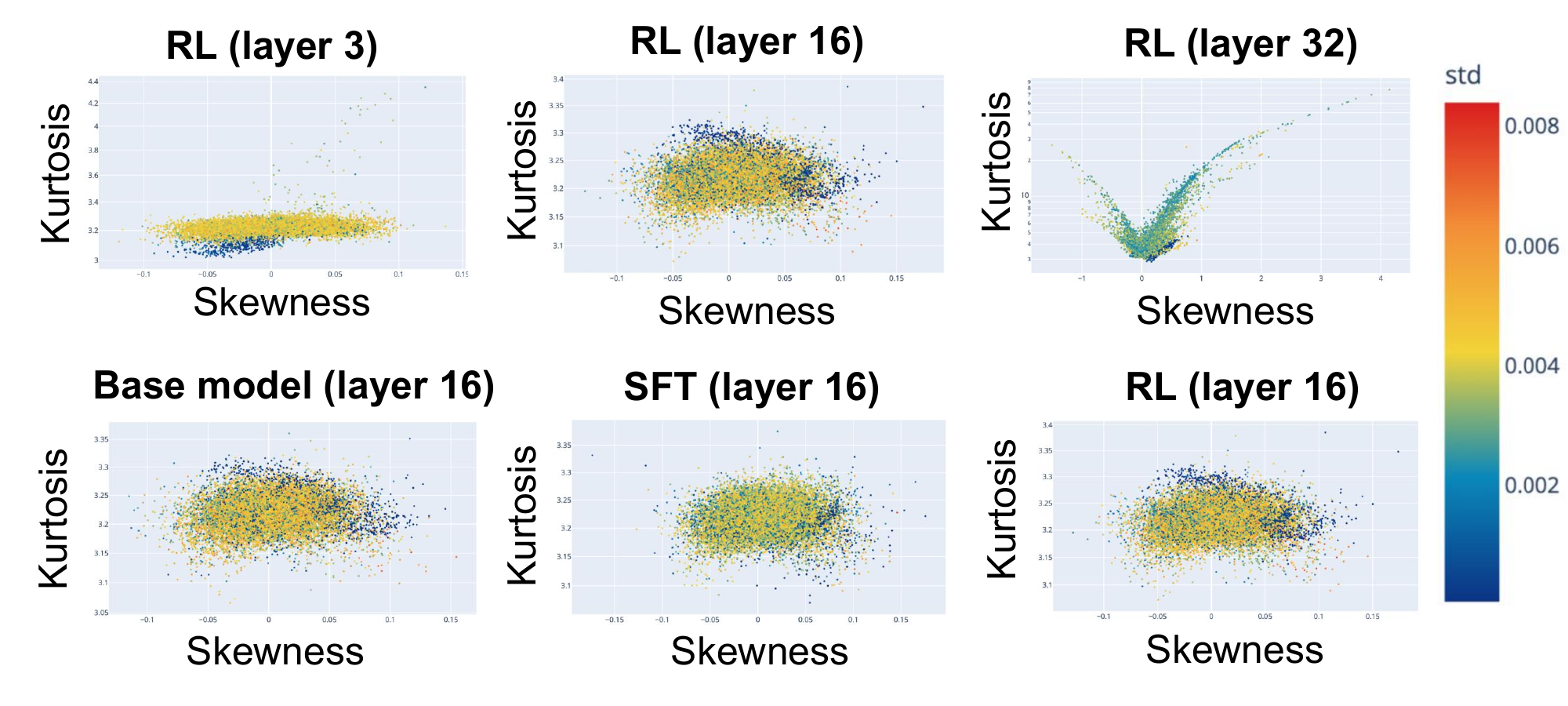}
  \caption{Skewness–kurtosis of logit weights across layers. RL shifts distributions from near-Gaussian (layer 3) to heavy-tailed, structured patterns (layer 32). At layer 16, base is compact and symmetric, SFT slightly broadens, while RL introduces more dispersed, structured variations.}
  \label{fig:distribution}
\end{figure}

\begin{figure*}[t]
\centering
\begin{subfigure}[b]{0.46\textwidth}
  \includegraphics[width=\linewidth]{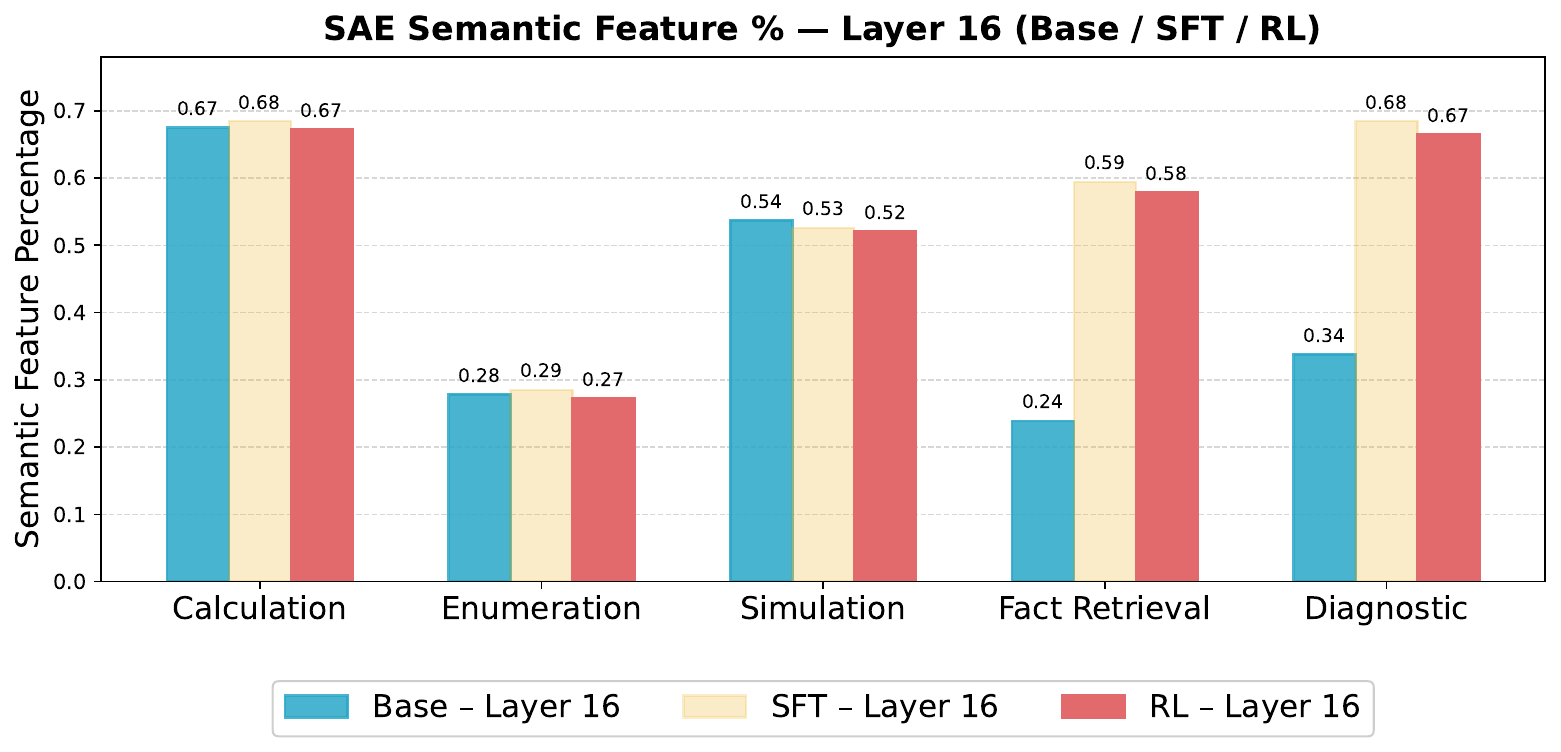}
  \caption{Different Strategy}
  \label{fig:sae_model}
\end{subfigure}
\hfill
\begin{subfigure}[b]{0.46\textwidth}
  \includegraphics[width=\linewidth]{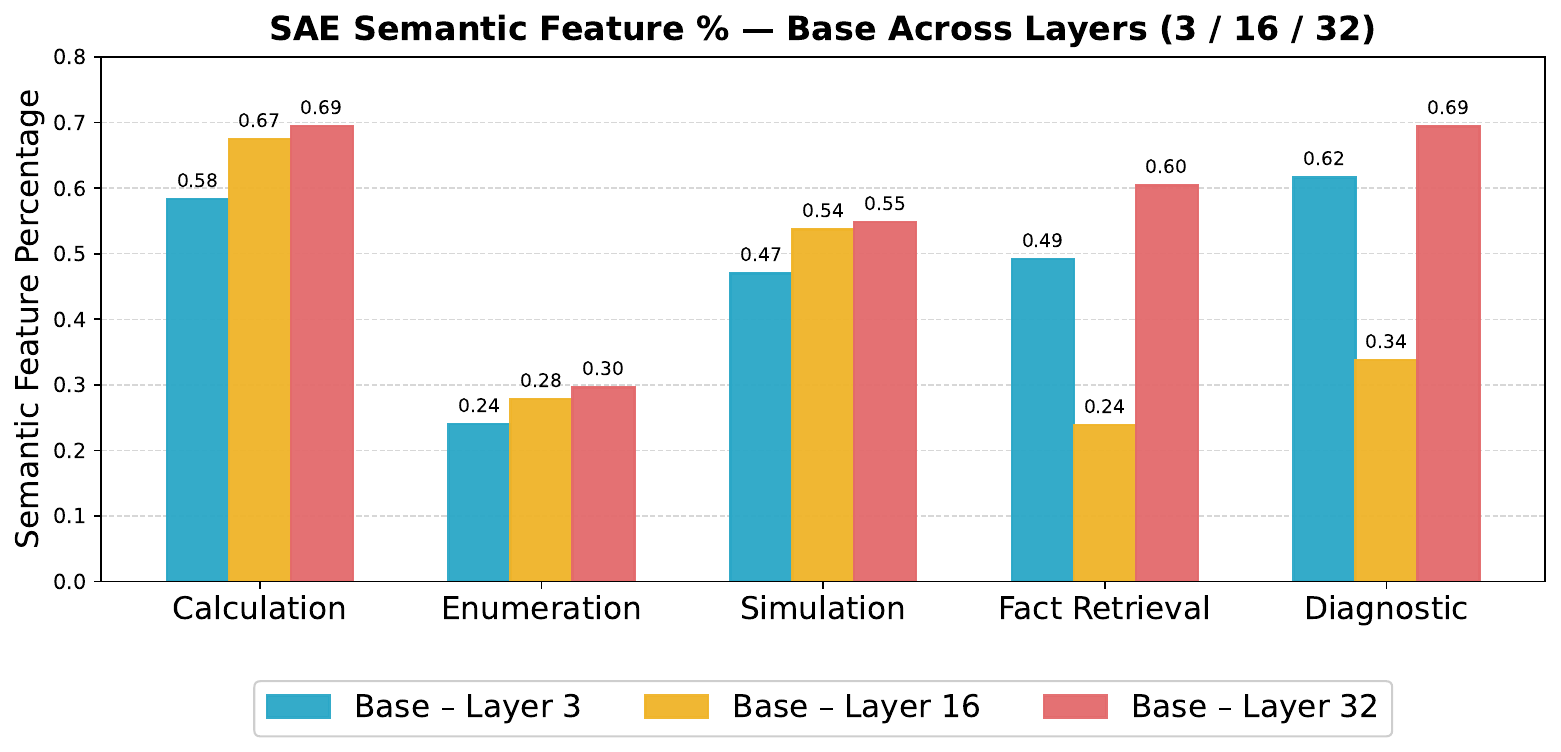}
  \caption{Different Layers}
  \label{fig:sae_layer}
\end{subfigure}
\caption{SAE-based cognitive-feature composition across strategies and layers.
We train sparse autoencoders (SAEs) on hidden states for different models (Base, SFT, RL) and layers, then compute the fraction of SAE features aligned with five cognitive behaviors: Calculation, Enumeration, Simulation, Fact Retrieval, and Diagnostic. Panel (a) compares post-training strategies at layer 16; panel (b) shows layer-wise trends for the Base model. 
}
\label{fig:sae}
\end{figure*}

\noindent\textbf{Interpretable subspaces tied to reasoning-related features.}
Figure~\ref{fig:sae} analyzes Qwen3-4B, Qwen3-4B-SFT, and Qwen3-4B-RL with sparse autoencoders (SAEs) to expose behavior-aligned subspaces. We fix depth at Layer 16 (the midpoint of Qwen3-4B) because mid-stack layers typically concentrate the richest, most compositional representations, providing a stable anchor for cross-model comparison. At this depth, both post-training strategies reallocate capacity toward knowledge access (Fact Retrieval) and self-verification (Diagnostic), while arithmetic (Calculation) and forward Simulation remain largely stable, and Enumeration stays sparse. Across depth in the base model, Calculation and Simulation strengthen gradually, whereas Fact Retrieval and Diagnostic trace a U-shaped trajectory, prominent in early and late layers with a mid-stack dip. Together, these patterns suggest a pipeline: early and deep layers focus on retrieval and checking, mid layers emphasize transformation and transport, and post-training amplifies retrieval and diagnostic subspaces without disrupting core arithmetic or simulation circuits.

\begin{table}[t]
\centering
\scalebox{0.8}{  
\small
\setlength{\tabcolsep}{3pt}
\begin{tabular}{lcccc}
\toprule
\textbf{Metric} & \textbf{SFT (no-think)} & \textbf{SFT (think)} & \textbf{RL} \\
\midrule
\multicolumn{4}{c}{\textit{Global parameter shift vs.\ Qwen3-14B base}} \\
\midrule
Changed parameters (\#)   & $1.44\times10^{9}$ & $1.44\times10^{9}$ & $1.44\times10^{9}$ \\
Change percentage (\%)    & 97.83              & 97.80              & 97.81              \\
Total change magnitude    & 29\,462.2          & 29\,033.8          & 29\,134.1          \\
Average change magnitude  & 66.51              & 65.54              & 65.77              \\
Maximum change magnitude  & 1\,544.5           & 1\,540.9           & 1\,556.3           \\
\midrule
\multicolumn{4}{c}{\textit{Distribution of change across major components}} \\
\midrule
Embed tokens change (\%)  & 97.04              & 97.04              & 97.03              \\
Layer-wise change (\%)    & 97.87              & 97.84              & 97.94              \\
Norm change (\%)          & 55.23              & 55.23              & 55.23              \\
LM-head change (\%)       & 97.89              & 97.88              & 97.94              \\
\bottomrule
\end{tabular}}
\caption{Parameter-space shifts of Qwen3-14B under SFT (no-think/think) and RL relative to the base. All methods update a similar share of parameters with comparable magnitudes, and changes are distributed similarly across embeddings, layers, norms, and the LM head, indicating that behavioral differences stem from training objectives rather than update scope.}
\label{tab:rl_param_shift}
\end{table}

\subsection{Low-Level Statistical Shifts and Surface Patterns}

\paragraph{Parameter-space shifts under post-training.}
To quantify post-training effects, we measure the fraction of parameters updated and their shift magnitudes for Qwen3-14B variants. As Table~\ref{tab:rl_param_shift} shows, roughly $98\%$ of parameters change in every variant, updates are pervasive, not sparse. RL’s total change magnitude is slightly below SFT (no-think) and comparable to SFT (think), indicating reward-based tuning does not necessarily induce larger global perturbations.

\paragraph{Component-level differences.}
Decomposing changes by module (Table~\ref{tab:rl_param_shift}) shows that embeddings and transformer layers account for most shifts across regimes, while normalization layers change much less ($55.23\%$). RL also exhibits a slightly larger maximum shift in the LM head than SFT (think), consistent with its objective of optimizing the output distribution via reward signals. This supports the view that RL primarily nudges the decision boundary by reshaping the output layer, whereas SFT distributes adjustments more uniformly across the network.

\section{Summary and implications for future study}

We presented a controlled benchmark that decomposes LLM reasoning into five core cognitive behaviors: calculation, enumeration, simulation, fact retrieval, and diagnostic checking, spanning four domains: mathematics, scientific reasoning, coding, and non-reasoning QA.  
Our three-stage pipeline enables reproducible, interpretable evaluation of how post-training regimes reshape these skills.
Our study reveals three key insights.  
First, RL-tuned models preserve a more balanced distribution of cognitive skills, supporting broader generalization.  
Second, SFT often induces over-specialization, improving narrow skills such as diagnostic, while weakening others like simulation.  
Third, these behavioral differences stem from the training objective rather than the parameter scale.

These results suggest that future reasoning-oriented LLMs should emphasize not only aggregate accuracy but also balanced skill development.  
Promising directions include behavior-aware objectives to prevent skill collapse, curriculum strategies to reinforce both domain-specific and transferable skills, and representation-level methods such as sparse autoencoders and activation steering to monitor and guide reasoning-related subspaces.
By revealing how post-training reshapes cognitive behaviors, our benchmark points toward training strategies that foster more robust, interpretable, and transferable reasoning capabilities.

\clearpage

% Bibliography entries for the entire Anthology, followed by custom entries
%\bibliography{anthology,custom}
% Custom bibliography entries only
%\bibliography{custom}
%\bibliographystyle{unsrt}   % or plain, ieeetr, abbrv, etc.

%\bibliographystyle{acl_natbib}
%\bibliographystyle{plainnat}

\bibliography{main}

\clearpage

\appendix

\section{Additional Experimental Details}~\label{app:expe-detail}

\subsection{Setup for Base, SFT, and RL Model Training and Evaluation}

\paragraph{Models.}
We adopt the recipe of \cite{huan2025does}: we (i) use their released Qwen3-14B SFT and RL checkpoints, (ii) replicate the procedure at 4B to obtain {Qwen3-4B} SFT/RL, and (iii) train {Qwen3-1.7B} SFT/RL while saving intermediate checkpoints for dynamics visualization.

Using {Verl} with GRPO and answer-correctness rewards for RL, we set: lr $=1\times10^{-6}$; global batch $=512$; clip $=0.22$–$0.28$; context $=16\mathrm{k}$ tokens; $16$ rollouts/prompt; mini-batch $=128$; KL/entropy coeffs $=0$; train $140$ steps (and use the corresponding checkpoint).
For SFT, we fine-tune with {LLaMA-Factory} on teacher CoT traces from {Qwen3-32B-Instruct} via reject sampling; lr $=5\times10^{-5}$; global batch $=512$; $1.5$ epochs (mirroring the RL horizon).
The dataset used is $47\mathrm{K}$ curated math problems, with CoT generated by {Qwen3-32B-Instruct}.
The baselines are {Qwen3-14B-Base} and {Qwen3-14B-Instruct} (evaluated in \emph{think}/\emph{no-think} modes).

\paragraph{Generation settings.}
Unless noted otherwise, we use nucleus sampling with temperature $0.6$ and top-$p=0.95$. For reasoning tasks, we cap generation at $32\,\mathrm{k}$ new tokens and apply stop sequences aligned with our standardized output headers. All decoding settings are held fixed across models for fair comparison.

\paragraph{Prompt template.} We use a single minimal prompt for all tasks:

\vspace{0.5cm}
\noindent\begin{minipage}{\columnwidth}
\begin{verbatim}
{problem}

Please reason step by step,
and put your final answer within \boxed{...}.
\end{verbatim}
\end{minipage}
\vspace{0.5cm}

Here, \texttt{\{problem\}} contains the task statement and, when applicable, any behavior-specific context (e.g., theorem tags for \emph{Guided} fact-retrieval or a lightly perturbed trace for diagnostic). Decoding settings are identical across models; evaluation reads only the final boxed answer.

\paragraph{Core metrics.}
Primary metric is \textbf{accuracy} under fixed decoding. We also compute behavior profiles (per-behavior accuracy vectors), normalized skill indices relative to the base model, and distributional drift of intermediate features when applicable.

\paragraph{Evaluation harness.}
We run batched inference with \texttt{vLLM} and a deterministic evaluation harness that records per-item seeds, prompts, responses, and scores. All artifacts (prompts, outputs, logs) are versioned and exported for audit.

\subsection{Setup for SAE Training and Interpretation}

\paragraph{Activation collection.}
We extract activations from specified transformer layers (e.g., blocks 3, 16, 32) at \texttt{hook\_mlp\_out} with context length 2048 over a balanced mixture of behavior-labeled prompts. We follow a stratified sampler to equalize behaviors and difficulties.

\paragraph{SAE objective \& launch.}
We train per-layer sparse autoencoders (SAEs) on transformer MLP outputs by minimizing
\(\mathcal{L}=\lVert h - D z\rVert_2^2 + \lambda \lVert z\rVert_1\),
where \(h\) is the hidden activation at \texttt{blocks.\(<L>\).hook\_mlp\_out}, \(z\) the sparse code, and \(D\) the decoder.
We use the open-source \texttt{sparsify} repo. Unless noted: \texttt{--ctx\_len 2048} and \texttt{--batch\_size 5}.

\noindent{Example command (single layer):}
\begin{lstlisting}[basicstyle=\ttfamily\scriptsize,breaklines=true,columns=fullflexible,keepspaces=true]
python -m sparsify <BASE_MODEL_NAME> <DATASET_OR_REPO> \
  --layers 16 --batch_size 5 --ctx_len 2048 \
  --data_preprocessing_num_proc 5 --k 250 \
  --run_name SAE-layer16-example
\end{lstlisting}

\noindent Replace \texttt{<BASE\_MODEL\_NAME>} (e.g., \texttt{Qwen3-4B-Base}) and \texttt{<DATASET\_OR\_REPO>} with model tag and dataset identifier or local path.

\paragraph{SAE interpretation with {Delphi}.}
For automated explanation of sparse autoencoder (SAE) and transcoder features, we use the open-source {Delphi} library \cite{paulo2024automatically}. {Delphi} programmatically \emph{generates} and \emph{scores} natural-language explanations for features, enabling large-scale interpretability; models can run locally or via OpenRouter. In our pipeline, for each learned feature we feed top-activating examples, obtain $k$ candidate explanations, score them with \textsc{Delphi}’s built-in scorer, and keep the top-ranked explanation subject to a quality threshold. 

\noindent{Example generation command:}
\begin{lstlisting}[basicstyle=\ttfamily\scriptsize,breaklines=true,columns=fullflexible,keepspaces=true]
python -m delphi <BASE_MODEL> <SAE_CHECKPOINT_DIR> \
  --hookpoints layers.16 \
  --scorers detection \
  --n_tokens 10_000_000 \
  --max_latents 10000 \
  --dataset_repo <DATASET_REPO> \
  --dataset_split 'train[:100%]' \
  --filter_bos \
  --name sae-layer16-explain
\end{lstlisting}
\noindent Replace placeholders with model tag, the directory containing the trained SAE for the target layer, and the dataset source. The flag \texttt{--hookpoints layers.16} selects the MLP-out features of layer 16 (other layers analogous); \texttt{--filter\_bos} removes BOS tokens; \texttt{--scorers detection} invokes the default explanation scorer.

\paragraph{Scoring SAE features for cognitive behaviors.}
After obtaining per-feature natural‐language explanations with \textsc{Delphi}, we assign scores for the five behaviors (calculation, enumeration, simulation, fact retrieval, diagnostic) using a lightweight LLM grader. For each behavior, we supply a behavior‐specific rubric (name, description, keywords, indicators) and batch a list of feature explanations; the model returns a JSON record per feature with a soft score, confidence, and a binary label. We decode \emph{greedily} (no sampling) to ensure schema validity and reproducibility. A feature is counted as “related” to a behavior if \texttt{score}\,$\ge 0.5$; we also report confidence–weighted variants (\(\tilde{s}=\texttt{score}\times\texttt{confidence}\)). Percentages are computed per layer and per model.

\noindent\textbf{Behavior‐parameterized prompt:}
\begin{lstlisting}[basicstyle=\ttfamily\footnotesize, breaklines=true]
You are analyzing feature explanations to determine if they relate to <BEHAVIOR>.

Category: <NAME>
Description: <DESCRIPTION>
Keywords: <KEYWORDS>
Indicators: <INDICATORS>

Analyze the following <N> features:
Feature 1 (ID: <feature_id_1>): <explanation_1>

...

For each feature, determine:
1) The extent to which it relates to <BEHAVIOR> (score 0.0-1.0).
2) Your confidence in this judgment (0.0-1.0).
3) A binary classification (0 or 1) for whether it relates to <BEHAVIOR>.

Respond with a JSON array (one element per feature) using EXACTLY:
[
  {"feature_id":"<feature_id_1>","score":<0.0-1.0>,
   "confidence":<0.0-1.0>,"classification":<0 or 1>},
  ...
]
Only return the JSON array.
\end{lstlisting}

\noindent\textit{Example instantiation (diagnostic).} Replace \texttt{<BEHAVIOR>} with \texttt{mathematical diagnostics} and populate \texttt{<NAME>, <DESCRIPTION>, <KEYWORDS>, <INDICATORS>} from our diagnostic rubric (Appendix~\ref{app:keywords}). The same template is used for the other four behaviors by swapping in the corresponding rubrics.

\paragraph{Interpretation.}
We align SAE latents to the five behaviors with a two–stage procedure. \emph{(i) Text-based alignment:} using \textsc{Delphi}, we generate per–feature explanations and grade them with a behavior-parameterized LLM prompt (shown above), which returns a soft score $s\!\in\![0,1]$, a confidence $\hat c\!\in\![0,1]$, and a binary label; a feature is considered related if $s\!\ge\!0.5$, and we also report confidence-weighted scores $s\cdot\hat c$. \emph{(ii) Signal-based validation:} we fit linear probes and compute mutual information between latent activations and behavior labels; any discrepancies trigger manual auditing of top-activating tokens/spans. We further summarize higher-order structure via skewness–kurtosis of decoder weights and visualize feature selectivity with activation-triggered exemplars, reporting layer-wise percentages for each model.

\subsection{Additional Details}
\noindent\textbf{Reproducibility and runtime.} We fix seeds for our experiments, and the evaluation harness, enable deterministic kernels where supported, and pin \texttt{vLLM} to a specific engine build. Experiments run on H100 (80\,GB) GPUs; inference uses tensor parallelism with static max sequence lengths, and batch sizes are tuned to avoid OOM while keeping GPU utilization $\ge 80\%$. We cache tokenizer outputs and KV states when permitted by the harness to ensure identical re-runs.

\paragraph{Limitations} Our analysis uses sparse autoencoders (SAEs) to link behavior-level signals to internal features; while informative, current SAEs techniques are compute-heavy and sensitive to design choices (e.g., hookpoints, sparsity targets). We therefore treat them as high-fidelity but non-definitive probes. A next step is to develop lightweight methods that still connect low-level statistics to hidden representations.

\paragraph{AI Assistance in Research and Writing.} This work uses AI tools for sentence-level proofreading.

\section{Cognitive Behaviors: Definitions, Keywords, and Indicators}\label{app:keywords}

Definition of the five cognitive behaviors: These behaviors represent fundamental modes of reasoning and problem solving—calculation, diagnosis, fact retrieval, simulation, and enumeration—each capturing a distinct way in which a system or learner processes information, manipulates symbols, and monitors or generates solutions across mathematical, scientific, and general domains.
%\clearpage

\begin{itemize}
\item \textbf{Calculation:} The execution of arithmetic, algebraic, or symbolic operations to derive quantitative or formal results from given inputs. Calculation emphasizes step-by-step manipulation under well-defined mathematical or logical rules, and may involve simplification, substitution, or transformation of expressions to reach a final result.

\textbf{Keywords:} compute, calculate, derive, simplify, evaluate, solve, result, step-by-step, formula, expression.

\textbf{Indicators:} explicit arithmetic/algebraic operations, transformations of equations, numerical outputs. 
\item \textbf{Enumeration:} The systematic and exhaustive generation of all elements, cases, or options satisfying a set of explicit constraints. Enumeration emphasizes coverage and completeness, often under conditions where solution space must be fully explored or cataloged for correctness, diversity, or comprehensiveness.

\textbf{Keywords:} list, all cases, options, possibilities, exhaustive, generate, under constraints, coverage.

\textbf{Indicators:} systematic listing, covering search space, combinatorial completeness.

\item \textbf{Simulation:} The mental or symbolic enactment of a process, system, or sequence of operations to predict or trace its behavior over time. Simulation requires stepwise reasoning consistent with formal rules (e.g., algebraic manipulation, causal transitions, or state-machine updates) and supports forecasting, testing, or illustrating dynamic evolution.

\textbf{Keywords:} simulate, step, next state, transition, apply rule, run through, dynamic, causal, evolve, predict.

\textbf{Indicators:} walking through process rules, state transitions, mental enactment of system behavior.

\item \textbf{Fact retrieval:} The recall or extraction of canonical, domain-grounded knowledge units (e.g., theorems, definitions, constants, statutes, or factual attributes). Fact retrieval emphasizes accuracy, grounding, and direct mapping to authoritative sources, rather than inference or elaboration.

\textbf{Keywords:} recall, definition, theorem, law, constant, fact, property, known, canonical, from memory.

\textbf{Indicators:} citing established knowledge, direct recall of named entities or rules without derivation.

\item \textbf{Diagnostic:} The act of identifying, localizing, and explaining the source of an error, inconsistency, or divergence in a reasoning process, system execution, or narrative account. In the context of LLMs, diagnostic reasoning is often accompanied by explicit self-monitoring or self-check language (e.g., “wait,” “let’s recheck,” “that step seems wrong”), which signals recognition of potential failure modes, causal attribution, and corrective reasoning.

\textbf{Keywords:} error, mistake, wrong, recheck, bug, contradiction, mismatch, correction, failure mode.

\textbf{Indicators:} self-check language (“wait,” “let’s recheck”), identifying divergence, explaining why an approach fails.
\end{itemize}

\subsection{Template and prompt for the dataset construction}

\paragraph{Candidate retrieval.}\label{app:prompt}
We retrieve candidates via dense \emph{embedding} search using \texttt{text-embedding-3-small}. Each item is indexed by the concatenation of its \emph{question} and \emph{reasoning trace}. For every behavior–domain seed, we compose a query from the seed summary, behavior keywords, and a few template sentences, then select the top-$k$ nearest items by embedding similarity (the model’s native similarity). We de-duplicate with an embedding-similarity threshold~$\tau$, retaining the highest-scoring item among any pairs above~$\tau$.

\noindent\textbf{Retrieval query template:}
\begin{lstlisting}[basicstyle=\ttfamily\footnotesize,breaklines=true,columns=fullflexible,keepspaces=true]
{SEED_SUMMARY}
Behavior: {BEHAVIOR} | Domain: {DOMAIN}
Keywords: {k1}, {k2}, {k3}
Template sentences: {TEMPLATE_SENTENCES}
(Find items semantically similar to this query using embedding similarity in
text-embedding-3-small.)
\end{lstlisting}

\subsection{Diagnostic Example}\label{app:diagnostic-example}

For the example in figure~\ref{fig:1.1.1}, starting from \(\sqrt{x+4}=x-2\), a first pass squares both sides to obtain \(x+4=(x-2)^2\), which simplifies to \(x^2-5x=0\) and yields candidate solutions \(x\in\{0,5\}\). A diagnostic self-check then enforces the original domain \(\sqrt{x+4}\ge 0\), giving \(x-2\ge 0\) and thus \(x\ge 2\), which rules out \(x=0\). Substituting into the original equation confirms the extraneous nature of \(x=0\) since \(\sqrt{4}=2\ne -2\). Therefore the only valid solution is \(x=5\). This sequence—derivation, reflection, constraint checking, and correction—illustrates diagnostic behavior by identifying an error introduced by a non-invertible operation (squaring) and revising the solution accordingly.

\subsection{Benchmark Scale and Extensibility.}
The benchmark spans \(5\times 4\) (behavior \(\times\) domain) cells, with each cell containing 50 to 200 carefully curated, high quality examples vetted for skill isolation, correctness, and calibrated difficulty. Seed questions are templatable and can be faithfully expanded, for example by renaming variables, rescaling parameters, swapping units or entities, or controlled paraphrasing, thereby increasing coverage without altering the targeted skill.

\section{Datasets Statistics}\label{app:data-distribution}

\begin{figure*}[t]
\centering
  \includegraphics[width=0.85\textwidth]{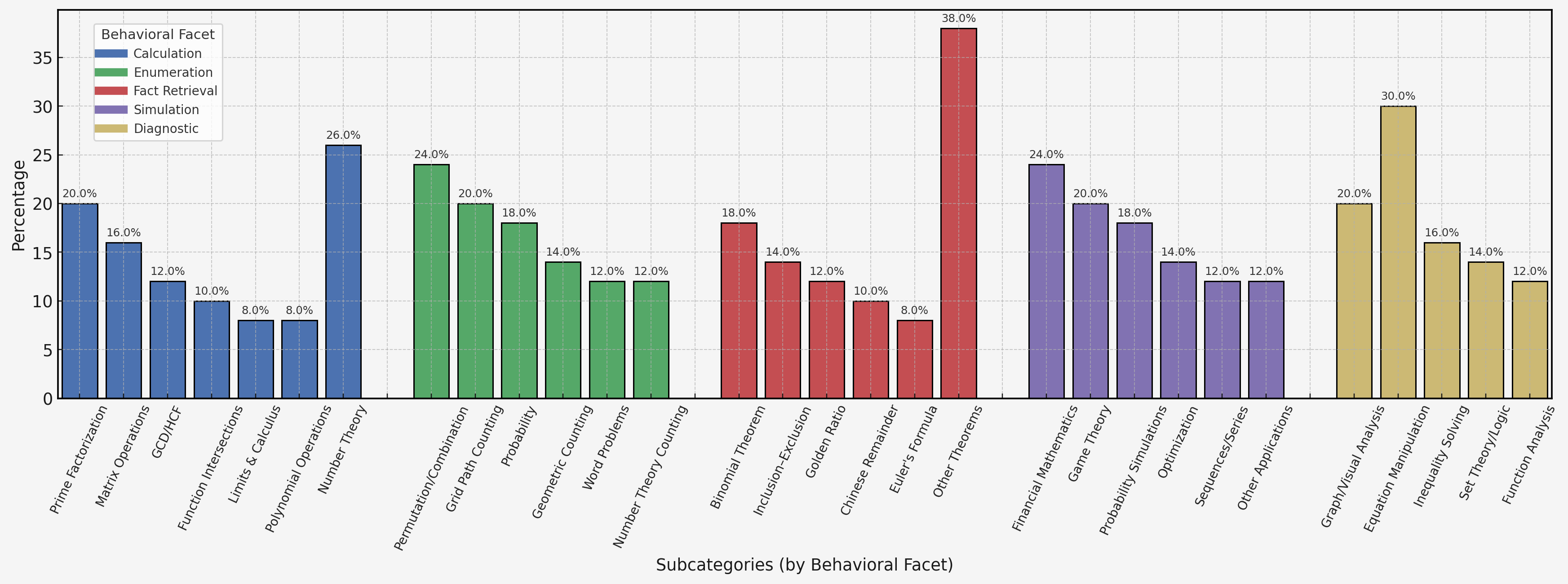}
  \caption{Distribution of math problems subcategories in our newly created dataset, grouped by behavioral facets: Calculation, Enumeration, Fact Retrieval, Simulation, and Diagnostic. Each bar shows the proportion (percentage) of problems belonging to a subcategory, with colored segments indicating the corresponding facet. This visualization highlights the relative prevalence of different cognitive skill types across the dataset.}
  \label{fig:math-stat}
\end{figure*}

\begin{figure*}[t]
\centering
  \includegraphics[width=0.85\textwidth]{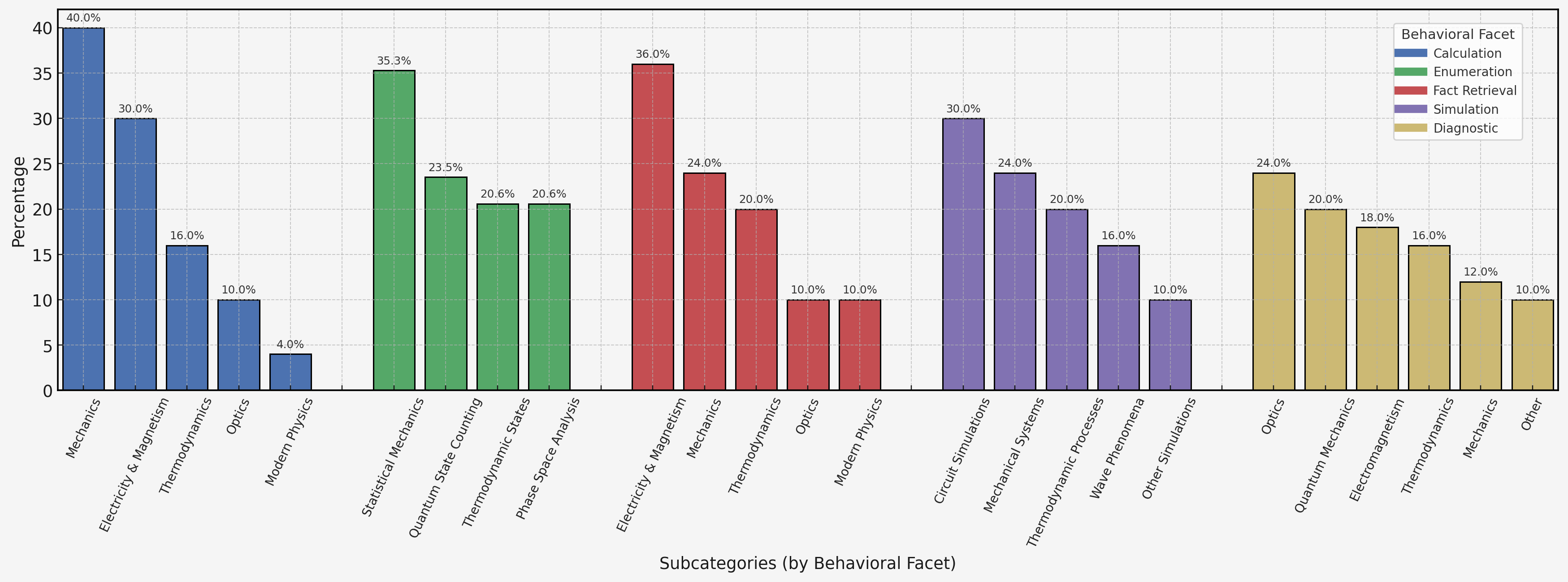}
  \caption{Distribution of physics subcategories in our newly created dataset, grouped by behavioral facets: Calculation, Enumeration, Fact Retrieval, Simulation, and Diagnostic. Each bar indicates the percentage of problems belonging to a specific subcategory, with colors corresponding to the facet type. The plot highlights the diverse coverage of physics domains such as Mechanics, Electromagnetism, Thermodynamics, Optics, and others across different cognitive skill requirements.}
  \label{fig:physics-stat}
\end{figure*}

\begin{figure*}[t]
\centering
  \includegraphics[width=0.85\textwidth]{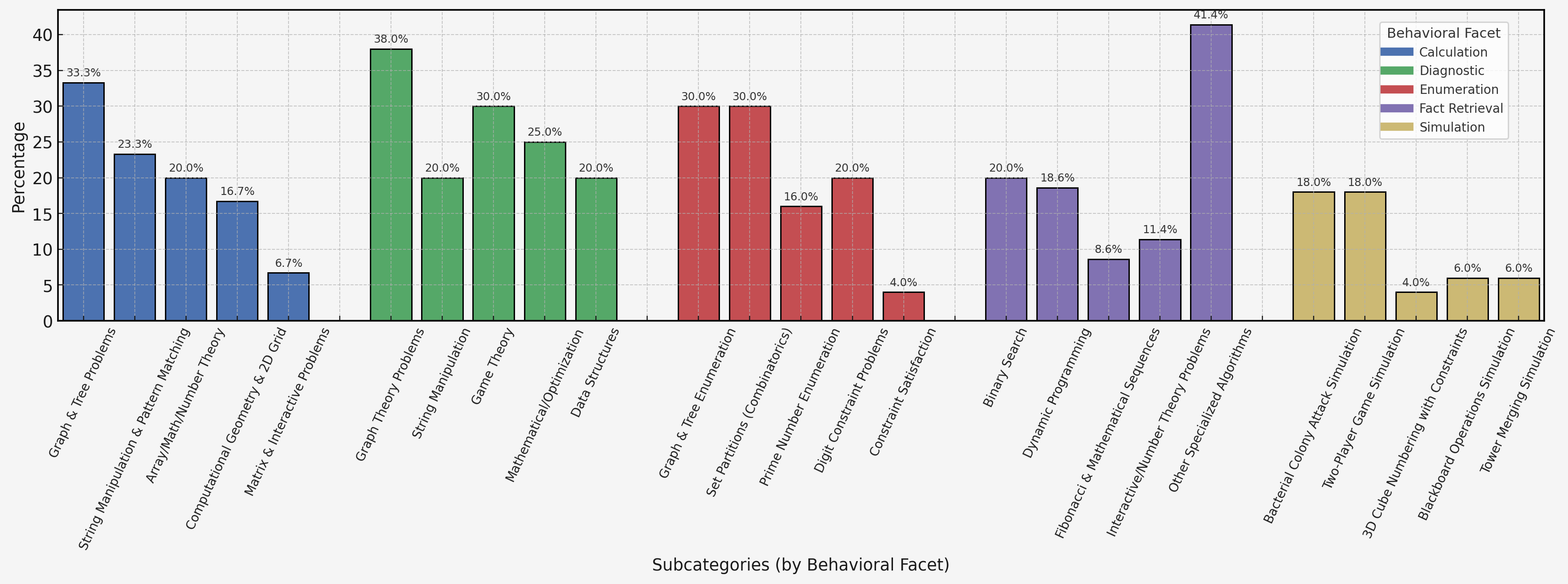}
  \caption{Distribution of code problem subcategories in our dataset, grouped by behavioral facets: Calculation, Diagnostic, Enumeration, Fact Retrieval, and Simulation. The figure highlights a diverse and well-balanced coverage of programming tasks including graph/tree problems, string manipulation, enumeration, dynamic programming, and simulation-based challenges, demonstrating the dataset’s quality and breadth for evaluating reasoning in algorithmic contexts. }
  \label{fig:code-stat}
\end{figure*}

\begin{figure*}[t]
\centering
  \includegraphics[width=0.85\textwidth]{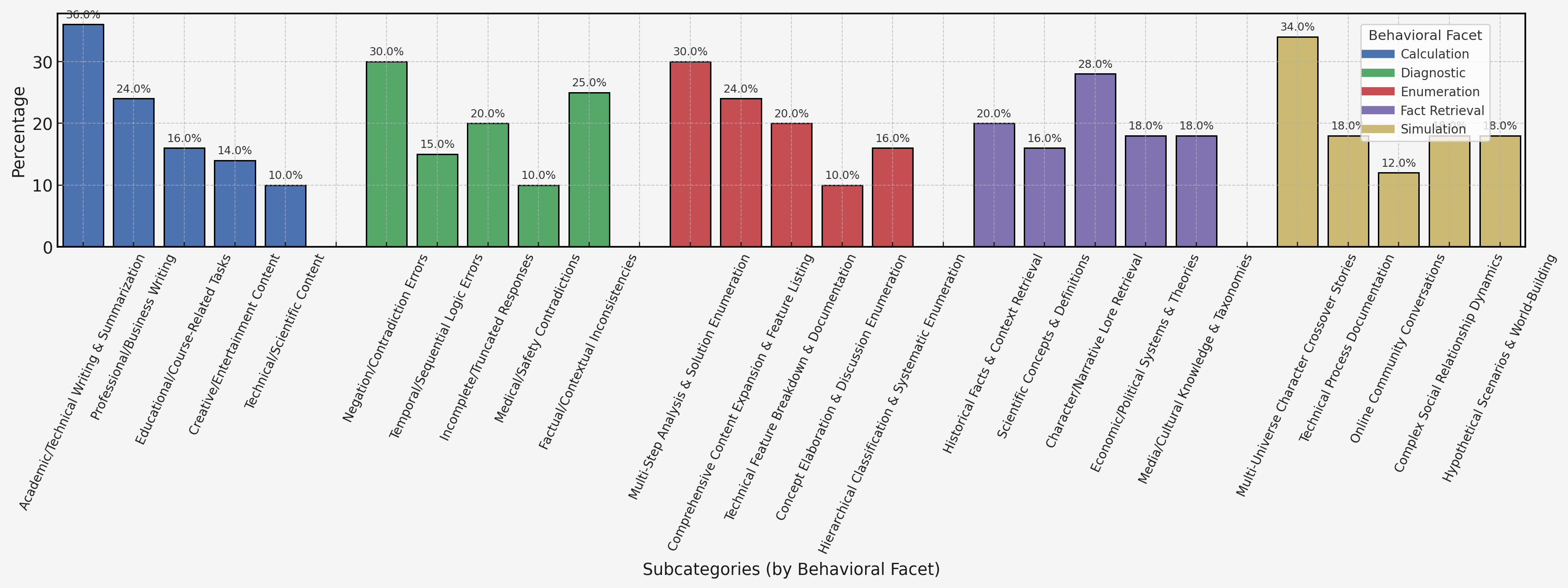}
  \caption{Distribution of non-reasoning subcategories in our dataset, grouped by behavioral facets: Calculation, Diagnostic, Enumeration, Fact Retrieval, and Simulation. The figure shows a diverse and balanced coverage of real-world tasks including writing and summarization, error detection, enumeration of content, knowledge retrieval, and interactive or scenario-based activities, underscoring the dataset’s breadth and quality for evaluating models beyond core reasoning skills. }
  \label{fig:non-reasoning-stat}
\end{figure*}

\noindent\textbf{Dataset statistics.}
Figures~\ref{fig:math-stat}, ~\ref{fig:physics-stat}, ~\ref{fig:code-stat}, and ~\ref{fig:non-reasoning-stat} present the distributions of subcategories across our newly created mathematics, scientific reasoning (physics), coding, and non-reasoning datasets, each further grouped by key behavioral facets: Calculation, Enumeration, Fact Retrieval, Simulation, and Diagnostic. The figures showcase a balanced and comprehensive coverage of both domain-specific and cross-domain skills, from core mathematical topics such as prime factorization, combinatorics, and binomial theorems, to scientific reasoning tasks in physics such as mechanics, thermodynamics, and optics, to algorithmic and computational challenges including graph/tree problems, dynamic programming, and simulation-based tasks, and finally to non-reasoning tasks such as writing and summarization, error detection, knowledge retrieval, and scenario-based simulations. This deliberate and systematic construction ensures that the dataset spans diverse problem types and cognitive demands, enabling robust evaluation of both reasoning-intensive and non-reasoning capabilities of large language models. The broad yet well-curated distributions highlight the quality, diversity, and representativeness of our benchmark, establishing it as a reliable resource for comprehensive assessment of models across reasoning, problem-solving, and real-world task performance.

\begin{table*}[ht]
\centering
\caption{\footnotesize Example problem templates across different mathematical domains. Each row shows a problem category, specific problem type, a template example with its solution, illustrating the diversity of mathematical reasoning tasks from basic arithmetic to error diagnosis. }
\label{tab:problem_templates}
\scalebox{0.8}{
\begin{tabular}{@{} l p{4cm} p{9.5cm} p{3cm} @{}}
\toprule
\textbf{Category} & \textbf{Problem Name} & \textbf{Template Example}  & \textbf{Answer}  \\
\midrule

\multirow{5}{*}{\textbf{Calculation}}
  &  Arithmetic gcd    & Calculate the greatest common divisor of 702 and 86814. & 234 \\
  & prime factors & Find the second-largest prime factor of 49766.  & 149 \\
  & arithmetic matrix eigenvalues    & Determine the largest eigenvalue by absolute value of the matrix: $[-8, 4, 4], [4, -1, 6], [4, 6, 5]$ & 10 \\
  & arithmetic matrix svd  & Calculate the rounded sum of all singular values from the SVD of: $[[3, -1, -3, 3], [-2, 2, 3, 4], [-2, 2, -3, 3], [-3, -4, 2, -1]]$  & 21 \\
  & algebra linear equation   & Solve $-523x + 5q + 30 = -522x$, $4x + 0x - 33q = 146$ for $x$. & 20 \\ 
\midrule

\multirow{5}{*}{\textbf{Enumeration}}
  &  combinatory distribution    & You have letters $\{'q': 4, 'l': 2, 'b': 2\}$. Distribute them into 3 labeled boxes of capacities $[4, 2, 2]$. How many ways? & 26 \\
  & logic gridworld & In a $7\times6$ grid, count all monotone paths from $(0,0)$ to $(6,5)$ avoiding cells $(6,0)$, $(3,5)$, $(1,1)$.  & 183 \\
  & geometric lattice enumeration    & Determine the largest eigenvalue by absolute value of the matrix: $[-8, 4, 4], [4, -1, 6], [4, 6, 5]$ & 1540 \\
  & combinatorial permutation constraint & Five couples sit in a row of 10 seats; no one sits next to their partner. How many seating arrangements are possible?  & 1263360 \\
  & geometric symmetry counting   & How many distinct ways to color the faces of a cube with 6  distinct colors so that adjacent faces have different colors? & 30 \\ 
\midrule

\multirow{5}{*}{\textbf{Simulation}}
  &   Number-sequence simulation  & The integers 1--120 are written on a board. Each minute replace each $n$ by $d(n + 3)$, the divisor-count function. After a day, what is the sum of the numbers on the board? & 240 \\
  & Stochastic walk / boundary-hitting & A frog starts at $(1, 2)$ inside a $4 \times 4$ square, jumping randomly by unit steps along coordinate axes until it hits the boundary. What is the probability it stops on a vertical side?  & $\frac{5}{8}$\\
  & Expected-value stopping time    & Joseph rolls a fair die repeatedly until he gets 3 identical consecutive rolls. What is the expected number of rolls? & 43 \\
  & Iterated redistribution process & Five balls are placed in the first five boxes of a row of six boxes. If any box has $\geq2$ balls, move one ball to the next box on the right. Repeat until all first five boxes have $\leq1$ ball. In how many initial placements does the last box end up empty? & 1296 \\
  & Financial iterative growth   & Bao invests $\$1000$ at $10\%$ annual compound interest. How much total interest is earned after 3 years? & 331 \\ 

\midrule

\multirow{5}{*}{\textbf{Fact Retrieval}}
&  Euler's formula / circle geometry  & Midpoint $H$ of chord $PQ$ in unit circle; chord length $2\sin\frac{t}{2}$. Find $H$ in terms of $t$. & $\left(\cos\frac{t}{2}\cos t, \cos\frac{t}{2}\sin t\right)$ \\
&Chinese Remainder Theorem / Euler's totient & Smallest $x \in \mathbb{N}$ with $x \equiv 3^{234} \pmod{700}$. & 169 \\
&Binomial/Chebyshev cosine multiple-angle & If $\cos\theta = \frac{1}{4}$, find $\cos 5\theta$. & $\frac{61}{64}$ \\
&Inclusion--Exclusion & Count integers $7 \leq n \leq 59$ relatively prime to 15. & 29 \\
&Great-circle distance (spherical law of cosines) & Surface distance on radius-$r$ sphere between $(p_1, q_1)$ and $(p_2, q_2)$. & $r\arccos(\sin q_1 \sin q_2 \cos(p_2 - p_1) + \cos q_1 \cos q_2)$ \\
\midrule

\multirow{5}{*}{\textbf{Diagnostic}}
&Problem-statement drift (sign/parameter flip) & Solver inadvertently changes the original problem (flips sign, replaces parameter), solving a different equation. \textbf{Example:} Asked to solve $x^2 - 3x + 2 = 0$ but solves $x^2 + 3x + 2 = 0$. & NA \\
&Internal inconsistency in trace & Derivation steps contradict stated givens or earlier results (e.g., variable values change mid-solution). \textbf{Example:} Declares $f(0) = 1$ initially but later uses $f(0) = 0$. & NA \\
&Tool / theorem mislabeling or misuse & Applies wrong theorem or uses one without meeting its conditions. \textbf{Example:} Uses Mean Value Theorem on a non-continuous or non-differentiable function. & NA \\
& Domain / legality mistakes & Ignores domain restrictions (division by zero, square roots of negatives, invalid trig ranges). \textbf{Example:} Cancels $(x - 1)$ without noting $x = 1$ is excluded. & NA \\
&Answer / format mismatch & Final answer doesn't match derived quantity or required format (units, boxed form, multiple-part). \textbf{Example:} Computes 12 but reports $3x^2 = 12$, or omits required components. & NA \\

\bottomrule
\end{tabular}}
\end{table*}\label{tab:math-examples}

\section{Detailed Description of the Meta-Evaluation Benchmark}

Table~\ref{tab:taxonomy} provides a comprehensive overview of meta-evaluation benchmark, which systematically categorizes evaluation tasks across 20 distinct problem types (4 domains $\times$ 5 cognitive behaviors). This structured framework enables granular assessment of LLM capabilities while maintaining consistency in evaluation methodology.

\subsection{Domain Categories}

The benchmark spans four complementary domains that represent core areas of AI capability assessment:

\textbf{Math Reasoning} encompasses problems requiring mathematical logic, symbolic manipulation, and quantitative reasoning. Tasks in this domain range from elementary arithmetic operations to advanced mathematical proof construction, testing models' ability to handle formal symbolic systems and rigorous logical deduction.

\textbf{Scientific Reasoning} evaluates understanding of physical principles, scientific methodologies, and domain-specific knowledge from fields such as physics, chemistry, and biology. These tasks assess whether models can apply scientific laws, perform calculations with physical units, and reason about real-world phenomena governed by natural principles.

\textbf{Coding} tests computational thinking, algorithm design, and programming proficiency. Problems in this category require models to understand data structures, optimize algorithms, debug code, and translate problem specifications into executable solutions across various programming paradigms.

\textbf{Non-reasoning} captures tasks that depend primarily on factual knowledge retrieval, procedural execution, or creative generation rather than complex logical inference. This category serves as a control to distinguish reasoning capabilities from memorization or pattern matching.

\subsection{Cognitive Behaviors}

The five cognitive behaviors represent fundamental modes of thinking that cut across domains:

\textbf{Calculation} involves executing well-defined computational procedures to derive quantitative or symbolic results. These tasks test precision, attention to detail, and mastery of algorithmic processes. Examples include computing prime factorizations, determining molecular radii from physical constants, implementing matrix operations, and calculating environmental metrics from model equations. The complexity varies from single-step arithmetic to multi-stage computations requiring careful tracking of intermediate results.

\textbf{Enumeration} requires systematic generation of all elements satisfying specified constraints. This behavior tests completeness, organizational thinking, and the ability to explore solution spaces exhaustively without omission or duplication. Tasks include combinatorial problems (such as distributing items into labeled boxes), generating quantum mechanical states (atomic orbital configurations), partitioning data structures (set partitions with ordering constraints), and cataloging instances (program features or article structures). Success demands both algorithmic rigor and verification that all cases have been considered.

\textbf{Simulation} involves mental or symbolic enactment of dynamic processes over time or through state transitions. These tasks evaluate sequential reasoning, state tracking, and the ability to project forward through cause-and-effect chains. Examples include iterating mathematical functions on a board, modeling physical processes (cooling or radioactive decay), tracing algorithmic execution (robot pathfinding or coin replacement dynamics), and predicting outcomes (weather forecasting or game simulation). The challenge lies in maintaining consistency across multiple steps while managing increasing complexity.

\textbf{Fact Retrieval} tests access to declarative knowledge and the ability to recall, recognize, or apply learned information. Tasks range from theorem retrieval in mathematics, recalling physical laws and formulas in science, accessing API documentation in coding, to retrieving cultural or procedural knowledge in non-reasoning contexts. This behavior assesses both the breadth of a model's knowledge base and its ability to retrieve relevant information when needed. Unlike pure memorization, many tasks require applying retrieved facts to novel situations.

\textbf{Diagnostic} evaluates meta-cognitive abilities, including error detection, consistency checking, and critical evaluation of reasoning processes. This highest-level cognitive behavior requires models to step back from problem-solving to assess the validity of solutions, identify logical flaws, or recognize ambiguities. Examples include verifying mathematical proofs, identifying errors in physical reasoning (such as perturbation analysis), debugging code, and detecting logical inconsistencies in arguments. Diagnostic tasks are particularly challenging because they require models to reason about reasoning itself.

\subsection{Problem Design Principles}

Each problem template follows several key design principles. First, problems are \textit{self-contained}, providing all necessary information within the problem statement to avoid ambiguity. Second, they have \textit{verifiable solutions}, enabling objective evaluation without subjective judgment. Third, they exhibit \textit{scalable difficulty}, allowing generation of instances ranging from simple to complex by adjusting parameters. Fourth, they are \textit{domain-representative}, reflecting authentic tasks that practitioners in each field actually encounter. Finally, problems are designed to be \textit{minimally ambiguous}, with clear success criteria that reduce evaluation uncertainty.

\subsection{Cross-Domain Patterns}

Several interesting patterns emerge from the taxonomy. Calculation tasks share common structure across domains---all involve applying defined procedures to inputs---but differ in the nature of operations (arithmetic vs.\ physical vs.\ algorithmic). Enumeration tasks universally require exhaustive search but vary in the combinatorial structure being explored. Simulation tasks all involve temporal or sequential progression but differ in whether the dynamics are deterministic or stochastic, discrete or continuous. Fact Retrieval varies most dramatically across domains, reflecting the distinct knowledge bases required. Diagnostic tasks converge on common meta-cognitive skills despite different surface manifestations.

\subsection{Use Cases}

This benchmark serves multiple purposes in AI evaluation research. It enables \textit{comparative analysis} of different models across specific cognitive dimensions, revealing whether a model excels at calculation but struggles with enumeration, for instance. It facilitates \textit{targeted improvement} by identifying precise capability gaps. It supports \textit{meta-evaluation} by providing a diverse testbed for assessing evaluation methods themselves---different evaluation approaches may perform better or worse across different cells of the taxonomy. Finally, it enables \textit{robustness testing} by generating multiple problem instances within each template to assess consistency of model performance.

\section{Additional Math Examples}\label{app:math-addition}

Table~\ref{tab:math-examples} illustrates representative problem templates for each behavior in mathematics. \emph{Calculation} covers core symbolic and numeric skills such as greatest common divisor, prime factors, eigenvalues and singular values of small matrices, and solving linear equations. \emph{Enumeration} includes combinatorial distributions, grid path counting, lattice enumeration, seating permutations with constraints, symmetry counting on polyhedra, and expression rationalization. \emph{Simulation} features stepwise processes such as number sequence updates, random walk boundary hitting, expected stopping time, iterative redistribution, and compound interest growth. \emph{Fact retrieval} targets canonically stated results (Euler’s circle geometry formulae, Chinese Remainder Theorem and totient, multiple–angle identities, inclusion–exclusion, spherical distance). \emph{Diagnostic} provides templated failure modes—including statement drift, inconsistent traces, theorem misuse, domain or legality mistakes, and answer or format mismatch—so models must detect and correct errors rather than compute alone. Each template admits numeric instantiations with a unique short answer, enabling scalable generation while preserving skill isolation.

\section{Training Dynamic}
\begin{figure*}[t]
\centering
\begin{subfigure}[b]{0.43\textwidth}
  \includegraphics[width=\linewidth]{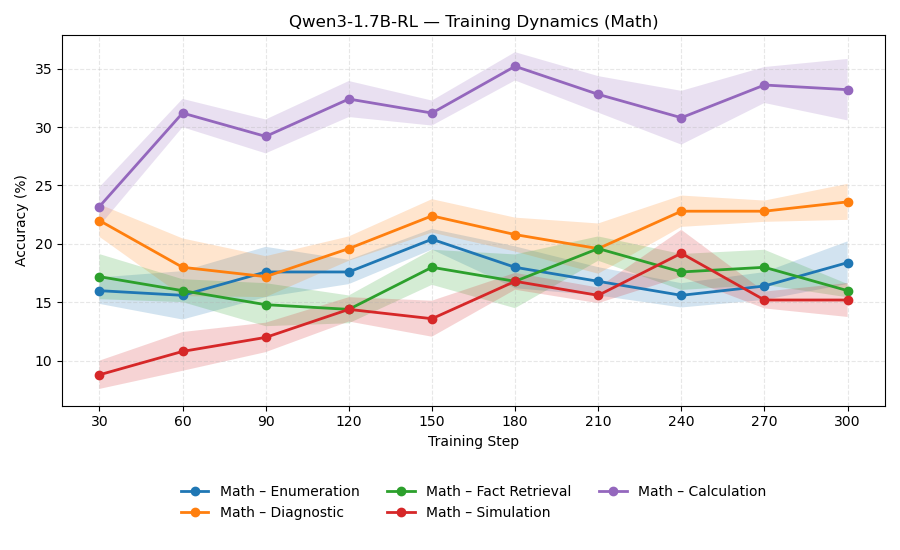}
  \caption{Math dynamic}
  \label{fig:sub1}
\end{subfigure}
\hfill
\begin{subfigure}[b]{0.43\textwidth}
  \includegraphics[width=\linewidth]{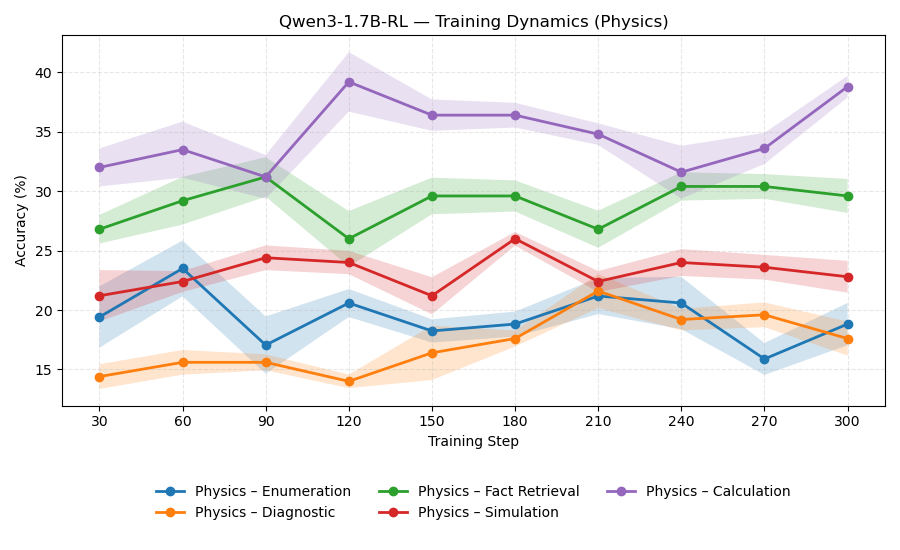}
  \caption{Physics dynamic}
  \label{fig:sub2}
\end{subfigure}
\caption{Training dynamics (Qwen3-1.7B-RL). Accuracy vs. step for math (a) and physics (b), with shaded variability. Calculation dominates; enumeration stays weakest/most volatile. Fact-retrieval, diagnostic, and simulation rise steadily (simulation relatively stronger in physics), indicating balanced RL gains rather than specialization.}
\label{fig:dynamic}
\end{figure*}

Under 1.7B RL, all five skills improve over training, with fast early gains that settle into a stable plateau. In {math}, {simulation} is consistently strongest, {diagnostic} rises steadily, and {calculation} shows clear improvement; {enumeration} and {fact retrieval} make smaller but persistent gains (see Fig.~\ref{fig:dynamic}). In {physics}, {calculation} leads throughout, {fact retrieval} strengthens over time, {simulation} improves to a moderate level, and {enumeration} and {diagnostic} advance more gradually. The shaded bands in Fig.~\ref{fig:dynamic} remain tight, indicating stable progress without regressions and balanced multi-skill benefits from RL.

\definecolor{lightyellow}{RGB}{250,237,215}
\definecolor{emphasisyellow}{RGB}{240,219,175}
\definecolor{slightlylesslightyellow}{RGB}{255,255,160}
\definecolor{lightblue}{RGB}{219, 238, 241}
\definecolor{slightlylesslightblue}{RGB}{160,160,255}
\definecolor{lightred}{RGB}{253, 202, 200}
\definecolor{slightlylesslightred}{RGB}{250, 150, 150}
\definecolor{lightgreen}{RGB}{235,250,210}
\definecolor{slightlylesslightgreen}{RGB}{160,255,160}
\definecolor{lightlightgray}{RGB}{230, 230, 230}

    \begin{tcolorbox}[colback=white,colframe=lightlightgray,fonttitle=\bfseries\large,colbacktitle=lightlightgray,enhanced,attach boxed title to top center={yshift=-10pt},title={\color{black}Math Diagnostic Example}]
        \vspace{10px}
        
        \begin{minipage}{\linewidth}
            \colorbox{emphasisyellow}{  
                \parbox{\dimexpr\linewidth-2\fboxsep}{
            \textbf{Problem:} Solve for all real \(x\):
\[
\sqrt{x+4} = x-2
\]
                }
            }

            \colorbox{lightgreen}{  
                \parbox{\dimexpr\linewidth-2\fboxsep}{
            \textbf{First Attempt:}   The square root is always non-negative, so \(x-2 \ge 0\), hence \(x \ge 2\). \\
  Squaring both sides:
  \[
  x+4 = (x-2)^2 \implies x+4 = x^2 -4x +4
  \]
  Simplify:
  \[
  0 = x^2 -5x \implies x(x-5) = 0
  \]
  Therefore, \(x = 0\) or \(x = 5\).
                }
            }

            \colorbox{lightblue}{  
                \parbox{\dimexpr\linewidth-2\fboxsep}{
            \textbf{Self-check/correction:}   \textbf{Wait, let’s re-check:} \\
  Since \(x \ge 2\), the candidate solution \(x = 0\) is invalid. \\
  Also, substituting \(x = 0\) into the original equation gives
  \[
  \sqrt{4} \neq -2
  \]
  so it does not satisfy the equation. \\
  Thus, the only valid solution is \(x = 5\).

                }
            }

            \colorbox{lightred}{  
                \parbox{\dimexpr\linewidth-2\fboxsep}{
            \textbf{Final Answer:} %\colorbox{slightlylesslightred}{Simulation} 
            
       \[
x = 5
\]
                }
            }

            \colorbox{slightlylesslightyellow}{  
                \parbox{\dimexpr\linewidth-2\fboxsep}{
            \textbf{Diagnostic:} 
            
            The diagnostic behavior is marked by the phrase “Wait, let’s re-check” (or similar), followed by detecting a domain or consistency issue and revising the solution.
                }
            }

        \end{minipage}
    \end{tcolorbox}
    %\caption{}
    \label{fig:1.1.1}

\clearpage

\end{document}